%% file: main.tex
\documentclass{article}

\usepackage{microtype}
\usepackage{graphicx}
\usepackage{subcaption}
\usepackage{booktabs} 
\usepackage{enumitem}
\usepackage{tabularx}

\usepackage{hyperref}

\usepackage[accepted]{icml2026}

\usepackage{amsmath}
\usepackage{amssymb}
\usepackage{mathtools}
\usepackage{microtype}
\usepackage{amsthm}
\usepackage[most]{tcolorbox}
\definecolor{PTGreen}{RGB}{92,134,110}
\definecolor{PTGreenDark}{RGB}{52,92,75}
\definecolor{PTLightGreen}{RGB}{242,250,245}
\definecolor{PTWhite}{RGB}{255,255,255}

\NewDocumentCommand{\zehong}
{ mO{} }{\textcolor{cyan}{\textsuperscript{\textit{Zehong}}\textsf{\textbf{\small[#1]}}}}

\usepackage[capitalize,noabbrev]{cleveref}

\theoremstyle{plain}

\theoremstyle{definition}

\theoremstyle{remark}

\usepackage[textsize=tiny]{todonotes}

\icmltitlerunning{Graph is a Substrate Across Data Modalities}

\begin{document}

\twocolumn[
  \icmltitle{Graph is a Substrate Across Data Modalities}



  \begin{icmlauthorlist}
    \icmlauthor{Ziming Li}{uconn}
    \icmlauthor{Xiaoming Wu}{nus}
    \icmlauthor{Zehong Wang}{nd}
    \icmlauthor{Jiazheng Li}{uconn}
    \icmlauthor{Yijun Tian}{nd} \\
    \icmlauthor{Jinhe Bi}{LMU}
    \icmlauthor{Yunpu Ma}{LMU}
    \icmlauthor{Yanfang Ye}{nd}
    \icmlauthor{Chuxu Zhang}{uconn}
  \end{icmlauthorlist}

  \icmlaffiliation{uconn}{University of Connecticut}
  \icmlaffiliation{nd}{University of Notre Dame}
  \icmlaffiliation{nus}{National University of Singapore}
  \icmlaffiliation{LMU}{LMU Munich}

  \icmlcorrespondingauthor{Chuxu Zhang}{chuxu.zhang@uconn.edu}

  \icmlkeywords{Machine Learning, ICML}
  \vskip 0.3in
]
\setlength{\floatsep}{2pt}
\setlength{\intextsep}{3pt}


\printAffiliationsAndNotice{}  
\begin{abstract}
Graphs provide a natural representation of relational structure that arises across diverse domains. 
Despite this ubiquity, graph structure is typically learned in a modality- and task-isolated manner, where graph representations are constructed within individual task contexts and discarded thereafter. 
As a result, structural regularities across modalities and tasks are
repeatedly reconstructed rather than accumulated at the level of intermediate graph representations. 
This motivates a representation-learning question: \emph{how should graph structure be organized so that it can persist and accumulate across heterogeneous modalities and tasks?} 
We adopt a representation-centric perspective in which graph structure is treated as a structural substrate that persists across learning contexts.
To instantiate this perspective, we propose \textbf{G-Substrate}, a \textbf{g}raph \textbf{substrate} framework that organizes learning around shared graph structures.
G-Substrate comprises two complementary mechanisms: a unified structural schema that ensures compatibility among graph representations across heterogeneous modalities and tasks, and an interleaved role-based training strategy that exposes the same graph structure to multiple functional roles during learning. 
Experiments across multiple domains, modalities, and tasks show that G-Substrate outperforms task-isolated and naive multi-task learning methods. The codebase, model, and datasets are available at \url{https://github.com/zmli6/G-Substrate}.
\end{abstract}

\input{sections/introduction}
\input{sections/shared_reasoning_substrate}
\input{sections/method}
\input{sections/experiment}

\input{sections/discussion}
\input{sections/related_work}

\input{sections/conclusion}

\section*{Acknowledgments}
This work was partially supported by the NSF under grants IIS-2528540, IIS-2334193, IIS-2340346, CNS-2426514, and CMMI-2146076. This work also used computational resources provided through NSF ACCESS grant CIS260048. Any opinions, findings, conclusions, or recommendations expressed in this material are those of the authors and do not necessarily reflect the views of the sponsors.


\section*{Impact Statement}

We propose a representation-centric framework that treats graph structure as a reusable intermediate substrate across tasks and modalities, with potential benefits for data efficiency and generalization in AI systems operating over structured information such as events, scenes, molecules, and algorithmic graphs.

A potential negative impact is that, because graph states are reused across tasks, systematic biases in how entities or relations are represented (e.g., in event corpora or visual datasets) may propagate across domains rather than remaining task-local. Responsible deployment therefore depends on careful dataset composition, transparent graph construction, and cross-domain evaluation. The present work also does not explicitly study how the composition of modalities, domains, or role types shapes representation formation; future work may explore principled strategies for balancing heterogeneous role-based supervision and extending this representation-centric principle to other forms of structured intermediate representations beyond graphs.

\bibliography{example_paper}
\bibliographystyle{icml2026}

\newpage
\appendix
\onecolumn
\input{sections/appendix}


\end{document}

%% file: sections/introduction.tex
\section{Introduction}
\begin{figure}[!ht]
  \centering
  \includegraphics[width=\linewidth]{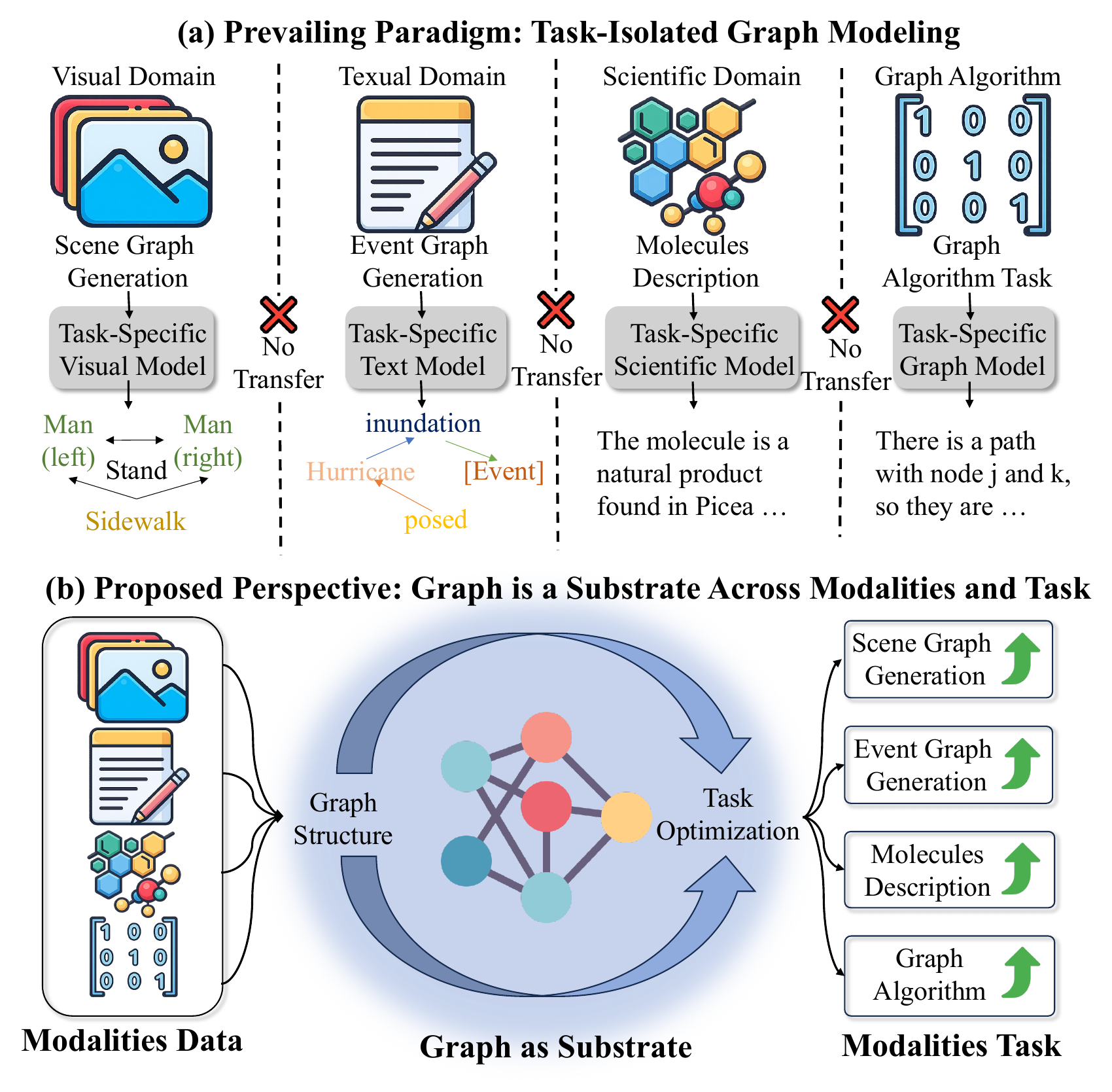}
\caption{
  \textbf{Task-isolated graph modeling vs. graph structure as a substrate.}
(a) Graph structure is learned in task-isolated pipelines, causing structurally similar graph patterns to occupy separate representation regions and limit cross-modal interaction.
(b) We organize graph structure as a shared substrate, encouraging graph patterns from different data modalities to converge and align, so structurally analogous configurations can mutually shape the representation and improve performance.
}
  \label{fig:framework}
  \vspace{-0.25in}
\end{figure}

Graphs provide a natural abstraction for relational information and arise across a wide range of domains and learning problems. For example, in computer vision, scene graphs encode objects and their interactions~\cite{ESGG:DBLP:conf/eccv/ChenWLZC24,FSGG:DBLP:conf/cvpr/LiZLC024}; in natural language processing, event graphs organize temporal and causal relations~\cite{hu-etal-2025-large,MAQInstruct:DBLP:conf/www/XuSZZ25}; in chemistry, molecular graphs represent atoms and bonds~\cite{Mol-LLaMA:DBLP:journals/corr/abs-2502-13449,GIT-Mol:DBLP:journals/cbm/LiuRTR24}; and in graph algorithmic tasks, graphs underlie tasks such as connectivity, shortest paths, and structural reasoning~\cite{GRAPH-R1:DBLP:journals/corr/abs-2508-20373,DBLP:journals/corr/abs-2509-24260}. 
Despite differences in input modalities and task objectives, graph structure provides an explicit structural interface through which heterogeneous data modalities can be organized
~\cite{DBLP:conf/icml/WangZMCZ025a}.
However, the widespread presence of graph-structured representations across tasks and modalities does not imply that learning systems are organized to preserve or accumulate graph structure. 
In practice, graph structure is built to serve a single objective and discarded after training, as shown in Figure~\ref{fig:framework}(a).
Many existing approaches instantiate structure as task-specific graph representations, such as supervision targets in scene graph generation~\cite{wu2025universal,R-SGG:DBLP:conf/aaai/LiuLW025} or event relation extraction~\cite{tao2025comprehensive,DBLP:journals/corr/abs-2508-20828}, despite structurally similar patterns recurring across these input modalities.
Recent efforts that aim to unify graph-centric learning largely focus on expanding task coverage or sharing model architectures, but still treat graph structure as a task-bound input or output rather than as a persistent intermediate state~\cite{DBLP:conf/naacl/SunMFMT25,DBLP:conf/nips/WangZCZ024,DBLP:conf/acl/WangWH00M24,DBLP:conf/nips/He0SC0LBH24}. These approaches are architecture-centric: they share model components, but do not establish a graph-level representation state that persists across tasks~\cite{DBLP:conf/icml/StandleyZCGMS20,DBLP:journals/corr/Ruder17a}. 
As a result, relational regularities found in one setting do not accumulate at the level of the graph, but remain confined to task-specific formulations. This exposes a representation-level mismatch: graph structure is treated as task-specific data rather than as a persistent learning state. 
The above issue motivates the following question:
\emph{how should graph structure be organized so that it can persist and accumulate across heterogeneous learning contexts rather than being reconstructed independently in each task?} Moreover, we do not attempt to unify task semantics but rather to align structural patterns that recur across domains.

There are two fundamental dimensions of heterogeneity that prevent graph structure from functioning as a reusable intermediate state, and each motivates a corresponding design requirement.
\textbf{Heterogeneity in form.} Graph structure varies widely across modalities and tasks in schema, granularity, and representational format (e.g., atom--bond triples in molecules vs.\ object--relation triples in scene graphs)~\cite{GraphOmni:DBLP:journals/corr/abs-2504-12764,chai2025graphllm}. This heterogeneity prevents direct reuse of graphs across learning contexts and motivates a \emph{structural compatibility} requirement: graphs from different contexts must be expressible in a common form so that they can coexist in a shared representation space.
\textbf{Heterogeneity in function.} Graph representations participate in learning under different functional roles: some tasks construct or refine graph structure, while others consume it for reasoning, prediction, or evaluation. A graph optimized under only one role becomes over-specialized to that role. This motivates a \emph{cross-role reuse} requirement: a reusable intermediate graph must remain functional under both structure-generate~\cite{DBLP:conf/icml/LaffertyMP01,DBLP:conf/cvpr/ZellersYTC18} and structure-understand roles~\cite{DBLP:journals/corr/abs-1806-01261,DBLP:conf/nips/HamiltonYL17}, so that representations are not over-fitted to a single objective.
Together, these two requirements directly motivate the two complementary mechanisms of G-Substrate: a unified structural schema (addressing form heterogeneity) and interleaved role-based training (addressing function heterogeneity).

Therefore, in this paper, we introduce a representation-centric view which considers \emph{graph structure as a persistent intermediate substrate} for coordinating learning across data modalities and functional roles, as shown in Figure.~\ref{fig:framework}(b). 
To operationalize this perspective, we introduce G-Substrate, a framework built around two complementary mechanisms: a \emph{unified structural schema} that establishes compatibility of graph representations across tasks and modalities, and \emph{interleaved role-based training} that exposes the same graph to multiple functional roles during learning. These mechanisms address structural and role heterogeneity, respectively.

We evaluate G-Substrate across tasks from multiple domains and modalities and show that it consistently outperforms task-isolated training and naive multi-task baselines. 
 Notably, the unified schema and role-based interleaving play complementary
roles: the schema yields gains once multiple tasks share the same graph state
space, and role-based interleaving further amplifies these gains by exposing
the same graph to multiple functional roles. Their combination consistently
yields the strongest performance, suggesting that the most robust graph
representations emerge when structural alignment is coupled with role-based
training.

%% file: sections/shared_reasoning_substrate.tex
\vspace{-0.1in}
\section{The G-Substrate Framework}
This section presents the G-Substrate framework and describes how the substrate-oriented perspective is realized in both data representation and model learning. Specifically, we formalize the central perspective of this work: \emph{graph structure as a persistent substrate rather than a task-bound artifact} (Section~\ref{sec:substrate}). This perspective leads to two design requirements, namely structural compatibility and cross-role reuse, which together define the design space of the framework.
We address the first requirement by organizing graphs within a unified graph
state space, aligning representations from heterogeneous tasks into a common
structural form (Section~\ref{sec:unified_schema}). We address the second requirement through interleaved role-based supervision, a training organization that exposes graph to multiple functional roles and promotes their reuse across learning contexts (Section~\ref{sec:interleaved}).

\vspace{-0.1in}
\subsection{Perspective: Graph is a Structural Substrate}
\label{sec:substrate}
 Graph structure arises across a wide range of learning problems, but is most often modeled in a task-bound manner. In many settings, such a structure is made explicit through graph representations. In prevailing practice, graph representations are constructed to serve individual task objectives and discarded thereafter, causing structural regularities that recur across tasks and modalities to be repeatedly reconstructed in isolation.
In contrast, we introduce a new representation-centric perspective: \textbf{\emph{graph is a reusable structural substrate across data modalities.}} Building on this perspective, we propose G-Substrate, a framework that organizes learning contexts across domains and modalities.

\begin{table}[h!]
	\centering
	\small
	\setlength{\tabcolsep}{5pt}
	\caption{
		Coarse topology statistics (per-graph averages).
		While global structural scale differs, recurring local structures are observed
		across all domains.
	}
	\vspace{-0.1in}
	\label{tab:sec2_structural_signal}
	\begin{tabular}{lcccc}
		\toprule
		Domain & AvgDeg & ASPL & TwoHop & Hubs \\
		\midrule
		Graph algorithm & 5.8 & 2.1 & 633 & 16 \\
		Molecular graphs & 2.1 & 6.2 & 53 & 12 \\
		Scene graphs & 1.5 & 1.4 & 2.4 & 0.7 \\
		Event graphs & 1.5 & 1.4 & 15 & 0.9 \\
		\bottomrule
	\end{tabular}
\end{table}

\begin{figure}[h!]
	\centering
	\includegraphics[width=\linewidth]{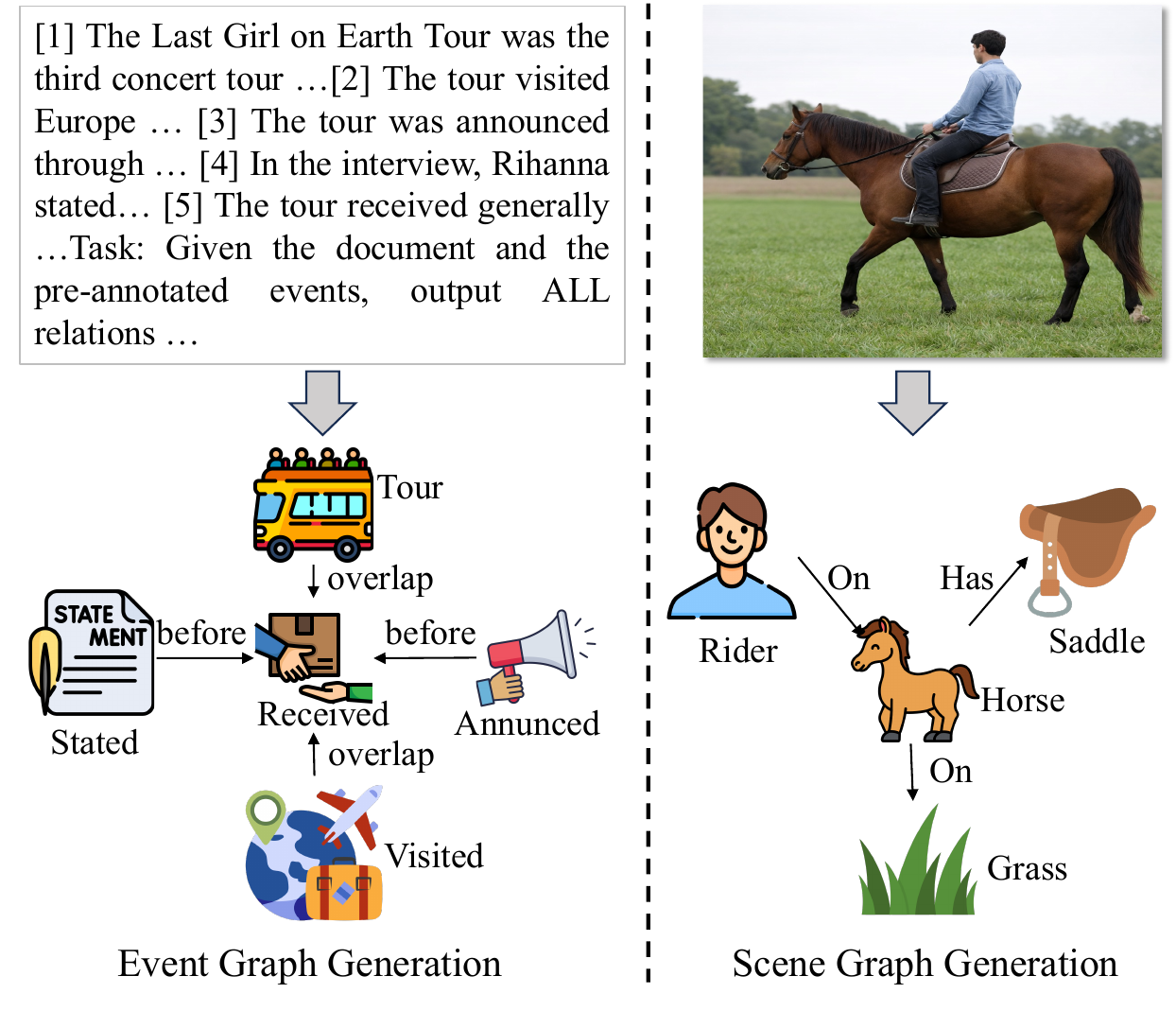}
	\vspace{-0.25in}
	\caption{
		Analogous constraint roles of hub motifs across tasks.
		In the event graph, the hub event \textit{received} participates in multiple
		temporal dependencies; in the scene graph, the hub object \textit{horse}
		participates in multiple spatial relations.
		The central node coordinates multiple relations and constrains
		their joint consistency.
	}
	\vspace{0.1in}
	\label{fig:motif_constraint_role}
\end{figure}

To empirically support this perspective, we examine whether structurally similar graph configurations recur across heterogeneous tasks and whether they play comparable structural roles despite differences in task semantics. We provide evidence for this perspective through quantitative structural statistics and qualitative motif analysis across heterogeneous domains, as reported in Table~\ref{tab:sec2_structural_signal}.  
These statistics summarize coarse topological properties, including average degree (AvgDeg), average shortest path length (ASPL), and the prevalence of simple local motifs such as two-hop chains and hub-centered patterns. Although global graph properties differ substantially, coarse local structures recur with non-trivial frequency in all settings studied. Beyond their prevalence, these graph structures play aligned functional roles.
Figure~\ref{fig:motif_constraint_role} shows a hub-centered configuration in an event graph and a scene graph. In the former, the event \textit{received} participates in multiple temporal dependencies; in the latter, the object \textit{horse} participates in multiple spatial relations. While task semantics differ, the central node in both cases coordinates multiple edges and constrains how they compose, indicating cross-domain invariance at the level of graph structure rather than task-specific meaning. These observations are drawn from representative datasets in scene graph generation, event relation extraction, molecular graphs, and algorithmic graph tasks, with detailed dataset descriptions and measurements provided in Appendix~\ref{app:empirical_motivation}.

To formalize this substrate-oriented view, we treat a graph as the fundamental structural representation. Specifically, a graph is defined as a set of structural triples
$
G = \{(u, r, v)\},
$
where \(u\) and \(v\) denote entities and \(r\) denotes a typed edge between them.
The relation label \(r\) and entity identities are preserved as part of the
structural representation; relations such as \emph{before} or \emph{wearing}
retain their inherent meaning. What this definition excludes is
\emph{task-specific framing}, such as loss functions, execution logic, and
optimization objectives, rather than the semantic content of relations and
entities themselves. A graph's identity is determined solely by the structural
configuration it encodes, not by how a particular task consumes it.
Entities or edges may carry optional attributes, which serve as auxiliary
annotations and leave the relational structure unchanged.

The graph substrate perspective treats graph as an intermediate structural representation intended to persist across learning contexts. For a graph to serve this role, two requirements follow. First, graphs arising from different learning settings must be structurally compatible so that they can reside in a unified representation space. Second, training must explicitly support the reuse of graphs across functional roles, rather than confining them to task-local roles.
G-Substrate operationalizes these requirements through a unified structural schema and an interleaved cross-task training strategy.

\begin{figure*}[!ht]
  \centering
  \includegraphics[width=\linewidth]{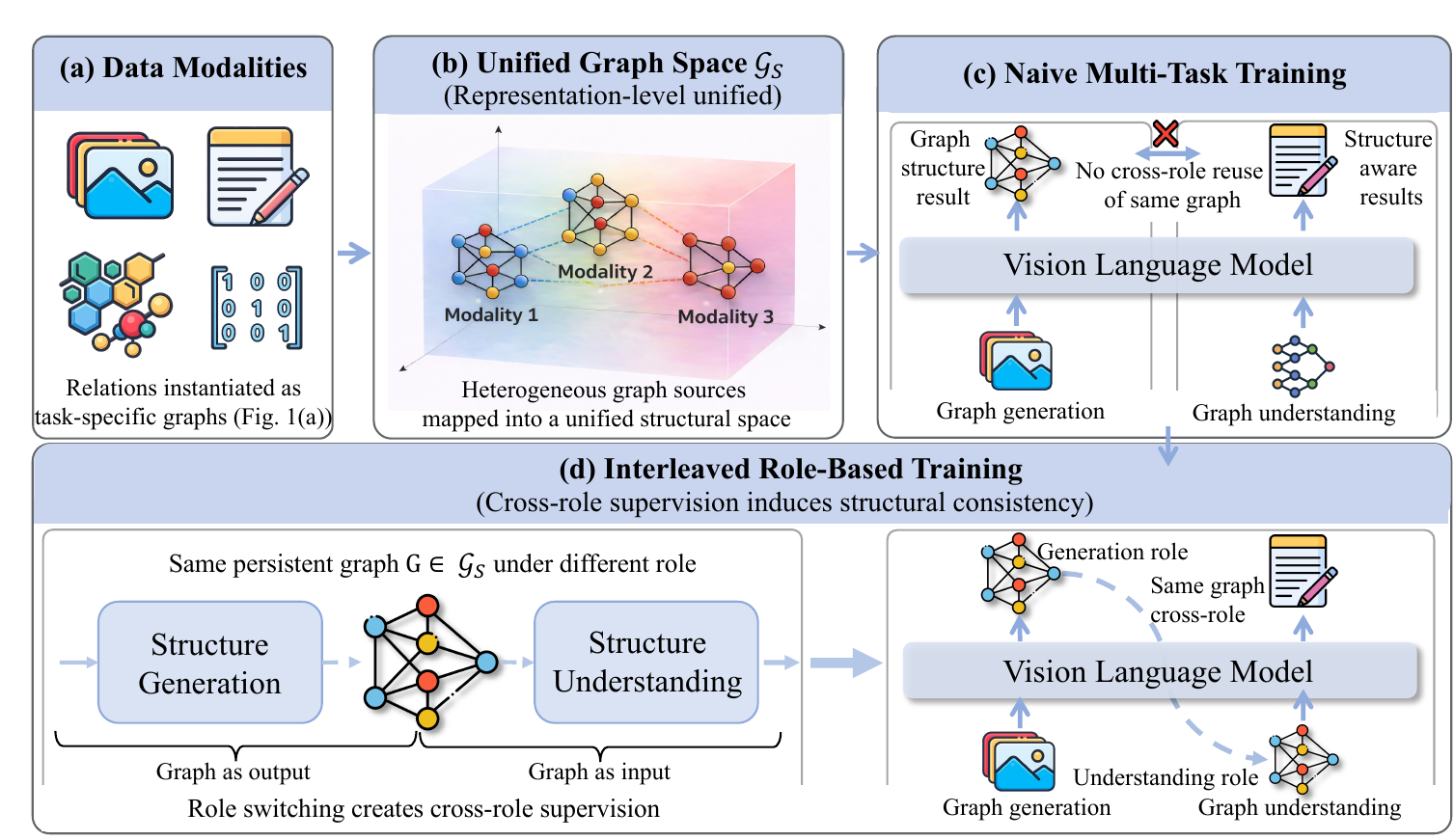}
  \vspace{-0.15in}
\caption{
\textbf{Unified graph substrate and cross-role training.}
Graph structures from heterogeneous modalities are mapped into a unified graph state space $\mathcal{G}_s$, where graphs serve as persistent structural representations (b).
Under naive multi-task training (c), graphs remain confined to fixed task roles, and the same graph is not reused across functional contexts.
Our interleaved role-based paradigm (d) exposes the \emph{same} graph $g \in \mathcal{G}_s$ to both structure-generation and structure-understanding roles, creating cross-role supervision.
This role switching induces structural consistency and supports reusable graph representations across tasks and modalities.
}

  \label{fig:method}
  \vspace{-10pt}
\end{figure*}

%% file: sections/method.tex
\vspace{-0.1in}
\subsection{Structural Compatibility: A Unified Schema}
\label{sec:unified_schema}

To ensure structural compatibility across tasks, G-Substrate organizes graphs
in a \emph{unified graph state space}. Building on the graph definition in
Section~\ref{sec:substrate}, we denote this space as
$\mathcal{G}_s = \{\, G \mid G = \{(u, r, v)\} \,\}$,
where each \(G \in \mathcal{G}_s\) consists of entities \(u, v\) connected by
typed edges \(r\).
Importantly, $\mathcal{G}_s$ is not the universal set of all conceivable
graphs, but a \emph{structured family} constrained by the unified schema: all
elements share consistent node identifiers, typed edges following fixed
conventions, and the same $(u, r, v)$ triplet format. Graphs that do not
conform to these conventions lie outside $\mathcal{G}_s$. This constraint is
what gives the space its utility as a shared representation: only by
restricting $\mathcal{G}_s$ to a structured family do graphs from heterogeneous
tasks become directly comparable and reusable.
Graphs arising from different modalities and tasks are mapped into this common structural
space, sharing consistent node identifiers, edge types, and connectivity rules.
Figure~\ref{fig:motif_constraint_role} gives examples of this mapping. An event graph
constructed from text
and a scene graph constructed from an image
are both represented as graph
instances \(G \in \mathcal{G}_s\). Although originating from different
modalities and tasks, these graphs share the same structural form, making hub-centered structural patterns (e.g., \textit{received} and
\textit{horse}) comparable in \(\mathcal{G}_s\).

Figure~\ref{fig:method}(b) provides a geometric intuition for this alignment.
Graphs from different modalities and tasks may initially occupy
disjoint regions under task-specific conventions. Expressing them in the unified
graph state space $\mathcal{G}_s$ brings these heterogeneous constructions
into a common structural region, where structurally compatible patterns become aligned.
Importantly, this concentration arises from explicit structural representation alignment
rather than from parameter sharing or feature similarity alone.

Structural compatibility alone, however, does not guarantee that graphs are
meaningfully exercised under diverse functional roles during learning. Even in
a shared structural space, a graph may still be optimized only in a single
usage context. Enabling unified graph representations to function consistently across roles
and tasks therefore requires an appropriate training organization, which is
provided by the interleaved role-based training described next.

\vspace{-0.1in}
\subsection{Cross-task Reuse: Interleaved Role-based Training}
\label{sec:interleaved}

A unified graph state space establishes structural compatibility of graphs across
tasks, but does not by itself determine how those graphs are used during
learning. Under naive multi-task training, different tasks are optimized jointly
using a shared backbone model, with each task receiving its native modality
input and producing task-specific outputs under its own objective. A common
instantiation of this naive setup is to use a single vision--language
foundation model as a shared backbone to process heterogeneous modalities,
with lightweight task-specific heads applied for different tasks~\cite{GITA:DBLP:conf/nips/WeiFJZZWK024,DBLP:conf/acl/Zhu0CXYZ25}. Although
graphs may implicitly arise in the shared model, they function primarily as
task-internal intermediates rather than as persistent representations reused
across tasks. As illustrated in Figure~\ref{fig:method}(c), graphs generated or
used in one task do not typically participate in other task roles or learning contexts. Consequently,
even when tasks operate in a common graph state space, their usage of graphs
remains largely isolated, resulting in limited cross-role interaction.

To enable graphs to function as persistent intermediate representations across
functional contexts, training must explicitly organize how they are reused under
different task types. Interleaved generation--understanding training pipeline provides
this organization. We model training as a sequence of \emph{task--role
instantiations} over unified graph states. Let $\mathcal{T} = \{T_1, \dots, T_K\}$
denote a set of tasks under different modalities, and $\mathcal{G}_s$ denotes the unified graph state
space. Each task $T_i$ is associated with a role function
\begin{equation}
\rho_i : \mathcal{G}_s \rightarrow \{\textsc{generate}, \textsc{understand}\},
\end{equation}
where \textsc{generate} corresponds to tasks that construct or refine the graph
structure (e.g., scene graph generation, event graph extraction), and
\textsc{understand} corresponds to tasks that operate on graph structure for
reasoning, prediction, or evaluation (e.g., graph algorithm).

Next, training is organized as a sequence
$\{(T_{i_t}, \rho_{i_t})\}_{t=1}^{N}$, in which graphs produced under generation
tasks may be reused as inputs under subsequent understanding tasks. To make the
input--output flow explicit, we view each task $T_i$ as an operator acting on
graphs and, optionally, modality inputs. Let $\mathcal{X}_i$ denote the
modality-specific input space associated with $T_i$, and let $\mathcal{Y}_i$
denote its task-specific output space. Each task induces a mapping
\begin{equation}
T_i : (\mathcal{X}_i, \mathcal{G}_s) \rightarrow (\mathcal{G}_s, \mathcal{Y}_i).
\end{equation}
When $\rho_i = \textsc{generate}$, the task produces or refines a graph:
\begin{equation}
G^{(t)} = T_i(x_i, G^{(t-1)}), \quad G^{(t)} \in \mathcal{G}_s.
\end{equation}
When $\rho_i = \textsc{understand}$, the task treats the graph as an
intermediate representation and produces predictions or supervision signals:
\begin{equation}
y_i^{(t)} = T_i(x_i, G^{(t)}), \quad y_i^{(t)} \in \mathcal{Y}_i.
\end{equation}
Interleaving therefore induces a trajectory of a graph.
$
G^{(0)} \rightarrow G^{(1)} \rightarrow \cdots \rightarrow G^{(N)},
$
where graphs serve as persistent intermediate representations that evolve
across successive generations and understanding tasks rather than being reconstructed
independently and discarded in each task, as illustrated in Figure~\ref{fig:method}(d).

\paragraph{Concrete example.}
Consider a two-step trajectory with role sequence
$(\textsc{generate}, \textsc{understand})$. At $t=0$, $G^{(0)}$ is the empty
graph. At $t=1$, given a street-scene image $x_1$, a scene graph generation
task produces $G^{(1)} =$ \{(\textit{rider}, \textit{on}, \textit{horse}),
(\textit{horse}, \textit{on}, \textit{grass}), (\textit{rider},
\textit{wearing}, \textit{helmet})\}.
At $t=2$, a graph reasoning task operates directly on $G^{(1)}$: given the
query \textit{``is there a path from \textbf{rider} to \textbf{grass}?''}, the
model traverses $G^{(1)}$ along $\textit{rider} \xrightarrow{\textit{on}}
\textit{horse} \xrightarrow{\textit{on}} \textit{grass}$ and returns
$y^{(2)} = \textit{yes}$. The graph $G^{(1)}$ itself is not modified. The same
graph state is thus produced under \textsc{generate} by a vision task and
immediately reused under \textsc{understand} by a structural reasoning task.

Schema compatibility extends this reuse beyond a single modality. For instance,
extracting typed temporal relations (e.g., \textit{before}, \textit{overlap})
from a news passage yields an event graph whose triplets share the $(u, r, v)$
format with $G^{(1)}$, so both reside in $\mathcal{G}_s$ despite differing
modalities and relation vocabularies.


From a representation-centric perspective, interleaving alters the supervision received by graphs rather than modifying individual task objectives. In task-isolated training, a graph is optimized under a single task type and only needs to satisfy constraints induced by that usage context. Under interleaved generation–understanding training, the same graph must remain usable across multiple task types. Graphs that support one type but are structurally incompatible with others receive inconsistent supervision and are gradually disfavored. This bias toward structurally coherent graphs emerges from the training organization itself, rather than from explicit regularization or parameter-level coupling.

%% file: sections/experiment.tex
\vspace{-0.1in}
\section{Experiments}
\label{sec:experiments}
\vspace{-0.05in}
This section examines whether organizing learning around reusable intermediate
graph leads to consistent improvements across heterogeneous learning
settings mediated by graph structure.
We study this question by contrasting G-Substrate with task-isolated training
and naive multi-task learning, and by conducting controlled analyses that
disentangle the roles of structural alignment and cross-role reuse of graph.
We additionally assess how a unified, substrate-oriented framework compares to
representative task-specific models under standard evaluation protocols.

\begin{table*}[t]
\centering
\small
\setlength{\tabcolsep}{5pt}
\caption{
\textbf{Main results across modalities, domains, and tasks.}
Best results are in \textbf{bold}; second-best are \underline{underlined}.
GAR (Graph Algorithmic Reasoning) is evaluated using \textbf{accuracy} for each task (CT: Connectivity, CD: Cycle Detection, SP: Shortest Path, BM: Bipartite Matching).
MGD (Molecular Graph Description) is evaluated using \textbf{BLEU-4} and \textbf{ROUGE-L}.
SGG (Scene Graph Generation) reports \textbf{PCIs R@50}.
ERE (Event Relation Extraction) reports \textbf{F1 scores} on MAVEN-S, MAVEN-T, MAVEN-C, and HiEvent.
}
\vspace{-0.1in}
\begin{tabular}{lcccc|cc|c|cccc}
\toprule

& \multicolumn{4}{c}{\textbf{GAR}} 
& \multicolumn{2}{c}{\textbf{MGD}} 
& \multicolumn{1}{c}{\textbf{SGG}} 
& \multicolumn{4}{c}{\textbf{ERE}} \\

\cmidrule(lr){2-5} \cmidrule(lr){6-7} \cmidrule(lr){8-8} \cmidrule(lr){9-12}

\textbf{Method}
& CT & CD & SP & BM
& BLEU-4 & ROUGE-L
& PCIs
& MA-S & MA-T & MA-C & HiE \\

\midrule
\multicolumn{12}{c}{\textbf{Task-Specific Training}} \\
\midrule

GITA~\cite{GITA:DBLP:conf/nips/WeiFJZZWK024}
& 98.17 & \textbf{98.07} & 39.15 & 93.19
& -- & --
& --
& -- & -- & -- & -- \\

G-Wiz~\cite{DBLP:conf/kdd/00010TL24}
& 97.74 & 95.46 & 41.46 & 92.15
& -- & --
& --
& -- & -- & -- & -- \\

M-LLama~\cite{Mol-LLaMA:DBLP:journals/corr/abs-2502-13449}
& -- & -- & -- & --
& \underline{50.74} & 67.02
& --
& -- & -- & -- & -- \\

PGSG~\cite{FSGG:DBLP:conf/cvpr/LiZLC024}
& -- & -- & -- & --
& -- & --
& \textbf{26.9}
& -- & -- & -- & -- \\

ProtoEM~\cite{DBLP:journals/corr/abs-2309-12892}
& -- & -- & -- & --
& -- & --
& --
& \underline{53.80} & 31.80 & 27.90 & 20.43 \\

LLMERE~\cite{hu-etal-2025-large}
& -- & -- & -- & --
& -- & --
& --
& \textbf{54.30} & 35.60 & 27.90 & \underline{22.90} \\

Naive single-task
& \underline{99.44} & 92.18 & 38.27 & 92.05
& 48.59 & 66.65
& 23.74
& 39.65 & \underline{41.60} & 27.70 & 17.10 \\

Unified single-task
& 97.80 & 94.70 & 37.14 & 85.98
& 47.35 & 65.64
& 22.43
& 45.45 & 33.29 & 30.22 & 14.28 \\

\midrule
\multicolumn{12}{c}{\textbf{Multi-Task Training}} \\
\midrule

Naive multi-task
& \textbf{99.71} & 94.72 & 41.27 & 92.21
& 48.11 & 66.11
& 24.68
& 36.87 & 39.14 & 37.02 &  18.78\\

Unified multi-task
& 98.09 & 96.19 & \underline{45.02} & \underline{94.23}
& 49.99 & \underline{67.36}
& 25.36
& 51.89 & 40.05 & \underline{40.75} & 19.37 \\

Naive multi-task + interleave
& 98.27 & 93.86 & 43.83 & 91.92
& 48.63 & 64.98
& 24.02
& 45.74 & 38.86 & 37.99 & 21.36 \\

\textbf{G-Substrate (Ours)}
& 98.41 & \underline{96.97} & \textbf{48.59} & \textbf{94.54}
& \textbf{51.53} & \textbf{68.47}
& \underline{25.38}
& 52.20 & \textbf{42.68} & \textbf{40.91} & \textbf{25.15} \\

\bottomrule
\end{tabular}

\label{tab:main_results}
\end{table*}

\vspace{-0.1in}
\subsection{Learning Settings and Tasks}
\vspace{-0.05in}
We evaluate the framework on four representative learning settings spanning
domains and modalities. For each task, we describe its objective, model
inputs and outputs, datasets, evaluation metrics, and task-specific baselines.

\textbf{Graph Algorithmic Reasoning (GAR).}
This task predicts the outputs of classical graph algorithms from an input
attributed graph. The model takes an attributed graph as input and outputs the
answer to a graph algorithmic query. We consider connectivity (CT), cycle
detection (CD), shortest path (SP), and bipartite matching (BM).
We follow the datasets and evaluation settings in prior work
~\cite{GITA:DBLP:conf/nips/WeiFJZZWK024,DBLP:conf/acl/WangWH00M24}, and report
accuracy as the evaluation metric. We compare against representative
task-specific models for graph algorithmic reasoning, including
GITA~\cite{GITA:DBLP:conf/nips/WeiFJZZWK024} and
GraphWiz~\cite{DBLP:conf/kdd/00010TL24}.

\textbf{Molecular Graph Description (MGD).}
This task requires generating a natural-language description of a molecule from
its structural representation. The model takes a molecular graph (atoms and
bonds), optionally accompanied by its SMILES string, as input and outputs a
textual description of molecular properties or functionality. We use the
Mol-Instructions dataset~\cite{Mol-Instructions:DBLP:conf/iclr/FangL0LH0FC24},
and evaluate using BLEU-4 and ROUGE-L. We compare against the task-specific
baseline Mol-LLaMA~\cite{Mol-LLaMA:DBLP:journals/corr/abs-2502-13449}.

\textbf{Scene Graph Generation (SGG).}
This task requires predicting a scene graph of objects and relations from an
input image. The model takes an image as input and outputs a structured graph
whose nodes correspond to objects and whose edges represent pairwise relations.
Evaluation is conducted on Visual Genome
~\cite{DBLP:journals/ijcv/KrishnaZGJHKCKL17} under the PCIs and SGCLs protocols,
reporting R@50 and mR@50, with PCIs R@50 as the primary metric. As ground-truth
bounding boxes are unavailable in our setting, we follow the data processing
protocol of~\cite{FSGG:DBLP:conf/cvpr/LiZLC024}. We compare against the
task-specific baseline PGSG~\cite{FSGG:DBLP:conf/cvpr/LiZLC024}.

\textbf{Event Relation Extraction (ERE).}
This task constructs event-relation graphs from text, capturing temporal,
causal, or subevent structures among events. The model takes raw text as input
and outputs a structured graph whose nodes correspond to events and whose edges
encode typed relations. We evaluate on MAVEN-subevent (MA-S), MAVEN-temporal
(MA-T), MAVEN-causal (MA-C)~\cite{DBLP:conf/emnlp/WangC0PWL00LLLZ22} and HiEvent
(HiE)~\cite{DBLP:conf/lrec/GlavasSMK14}, reporting precision, recall, and F1
score. We compare against task-specific baselines ProtoEM~\cite{DBLP:journals/corr/abs-2309-12892} and LLMERE~\cite{hu-etal-2025-large}.




\subsection{Training Paradigms}
\label{sec:exp_paradigms}
\vspace{-0.05in}

We compare learning settings that differ along two orthogonal axes:
(1) the representation of graph (task-specific vs.\ unified structural
schema), and
(2) the training organization (task-isolated, jointly multi-task, or with
interleaved role-based training).
This leads to six paradigms.
\textbf{Naive single-task (NST)} and \textbf{Unified single-task (UST)} train
each task in isolation, differing only in whether graph uses native
formats or the unified schema.
\textbf{Naive multi-task (NMT)} and \textbf{Unified multi-task (UMT)} jointly
train all tasks, again differing in representation format but without exposing
the same graph to multiple functional roles.
\textbf{Naive multi-task + interleave (NMT-I)} introduces role-based interleaving
on top of naive task-specific representations, allowing the graph to be
reused under different task roles without structural alignment.
\textbf{G-Substrate (Unified + interleave, G-Sub)} combines the unified schema
with interleaved role-based training.
Together, these paradigms disentangle the effects of structural alignment and
cross-role reuse. Detailed definitions are given in
Appendix~\ref{app:training_paradigms}.
All methods share the same backbone model and optimization settings.
For multi-task settings, no additional task-specific fine-tuning after
training or test-time adaptation is applied; each model is trained once under
its corresponding paradigm and evaluated directly.
Unless otherwise specified, experiments use the Qwen3-VL-2B-Instruct
model~\cite{qwen3technicalreport} as the backbone.
Detailed training configurations are provided in Appendix~\ref{app:hyperparams}.

\subsection{Main Results}
\label{sec:exp_main}
Table~\ref{tab:main_results} summarizes the main results. Although G-Substrate uses a single unified model rather than domain-specialized architectures, it matches or exceeds task-specific systems on most metrics. On GAR, SP rises from G-Wiz's 41.46 to 48.59. On MGD, it reaches 51.53 BLEU-4 and 68.47 ROUGE-L, exceeding M-LLaMA's 50.74 and 67.02. On ERE, F1 on MA-T, MA-C, and HiE rises from LLMERE's 35.60, 27.90, and 22.90 to 42.68, 40.91, and 25.15. The gains are largest where evaluation rewards relational reasoning rather than local pattern matching, such as SP, BM, and multi-hop event relations, and smallest in structurally compact settings, where PGSG still leads SGG, 26.9 to 25.38, and LLMERE retains a narrow edge on MA-S, 54.30 to 52.20. We interpret this as evidence that organizing learning around a shared substrate carries enough structural inductive bias to match domain-specialized pipelines without sacrificing per-domain capability, while leaving room for task-specific tuning where graphs are small enough that specialization itself is the dominant lever.

We next analyze the effect of different training paradigms. G-Substrate
outperforms both task-isolated training and naive multi-task learning on most
metrics. The improvements are more pronounced in settings with stronger
structural demands, suggesting that the gains are tied to structural reasoning
rather than uniform scaling effects.
These patterns are consistent with the intended mechanism of G-Substrate.
Task-isolated training restricts graphs to a single functional context, while
naive multi-task learning, despite parameter sharing, does not require the same
graph to remain usable across roles. By contrast, G-Substrate combines
structural alignment with interleaved generation--understanding training,
encouraging graphs to remain valid under multiple roles. This cross-role
pressure biases representations toward relational regularities rather than
task-specific shortcuts, aligning with the observed performance trends.  Detailed results are provided in
Appendix~\ref{app:full_results}. We
additionally compare against the gradient-balancing multi-task baseline
GradNorm in Appendix~\ref{app:gradnorm}, where G-Substrate outperforms
NMT+GradNorm across all four domains without any explicit loss reweighting,
indicating that the dominant bottleneck is representational rather than
optimization-level.



\begin{table*}[!t]
\centering
\small
\caption{
\textbf{Effect of Schema Realization.}
Performance comparison of different schema realizations under identical multi-task training conditions.
The best-performing method is shown in \textbf{bold}.
}
\vspace{-0.1in}
\begin{tabular*}{\textwidth}{@{\extracolsep{\fill}} lcccc|cc|c|cccc @{}}
\toprule
\textbf{Method}
& \multicolumn{4}{c}{\textbf{GAR}} 
& \multicolumn{2}{c}{\textbf{MGD}} 
& \multicolumn{1}{c}{\textbf{SGG}} 
& \multicolumn{4}{c}{\textbf{ERE}} \\

\cmidrule(lr){2-5} \cmidrule(lr){6-7} \cmidrule(lr){8-8} \cmidrule(lr){9-12}

& CT & CD & SP & BM
& BLEU-4 & ROUGE-L
& PCIs
& MA-S & MA-T & MA-C & HiE \\
\midrule

Natural Language
& 97.15 & 94.20 & 44.80 & 92.75
& 49.20 & 66.80
& 24.10
& 50.80 & 40.95 & 39.20 & 23.90 \\

XML-style
& 94.80 & 92.30 & 40.10 & 88.40
& 44.60 & 60.50
& 23.65
& 46.30 & 36.40 & 34.70 & 20.80 \\

\textbf{Ours}
& \textbf{98.41} & \textbf{96.97} & \textbf{48.59} & \textbf{94.54}
& \textbf{51.53} & \textbf{68.47}
& \textbf{25.38}
& \textbf{52.20} & \textbf{42.68} & \textbf{40.91} & \textbf{25.15} \\

\bottomrule
\end{tabular*}

\label{tab:exp_schema}
\end{table*}

\subsection{Analysis}
We conduct controlled studies to analyze the mechanisms underlying
G-Substrate, isolating representation and training-organization factors while
keeping the backbone, data, and training budget fixed. Specifically, we examine:
(i) the interaction between structural alignment and role-based training,
(ii) the effect of schema realization,
(iii) cross-domain structural transfer,
(iv) the contribution of different cross-role training instantiations,
(v) the role of structural correctness of the reused graph, and
(vi) the impact of the proportion of role-based interleaving.
\subsubsection{Unified Strategy Analysis}
\label{sec:exp_strategy}
\textbf{Schema--Training Interaction.}
We examine whether the effect of unified representations arises from the
structural schema itself or from its interaction with role-based training.
Table~\ref{tab:main_results} shows that the \textit{Unified Single-Task} setting
does not outperform the \textit{Naive Single-Task} baseline and often performs
worse under task-isolated training. By contrast, in the multi-task setting the
unified schema yields consistent improvements over its naive counterpart
(\textit{Unified Multi-Task} vs.\ \textit{Naive Multi-Task}), and these gains
are further amplified once the same graph is exposed to multiple functional
roles during training. This indicates that the schema primarily establishes
structural compatibility, whose benefits emerge once graphs are shared across
tasks and grow strongest under role-based reuse.

\textbf{Effect of Schema Realization.}
We compare different realizations of the unified schema, including
natural-language descriptions, XML-style serializations, and the schema
representation used in G-Substrate, all encoding an identical graph under
the same role-based training setting. Table~\ref{tab:exp_schema} shows that
although alternative serializations permit basic transfer, their performance is generally
less stable. XML-style formats, in particular, tend to underperform, likely
because strict formatting encourages attention to surface structure rather than underlying 
relational semantics. The proposed schema realization provides more reliable
performance, indicating that effective structural reuse depends not only on
schema unification, but also on how relational structure is expressed when the graph is exercised under multiple functional roles during training.

\textbf{Cross-domain Structural Transfer.}
To assess cross-domain reuse of graph structure, we transfer from
event-centric text graphs to scene graph generation.
Table~\ref{tab:cross_domain_transfer} reports performance relative to a base
model without domain-specific pretraining.
Training on event graphs alone improves scene graph generation despite the
absence of target-domain supervision.
This suggests that learning organized around an explicit graph structure can
capture structural regularities that transfer across domains, rather than being
fully tied to a single task or modality.

\begin{table}[!t]
\centering
\small
\caption{
\textbf{Cross-domain structural transfer from event graphs to scene graph generation.}
Models are evaluated on scene graph generation (PCIs R@50).
$\Delta$ denotes the absolute performance change relative to the Base model.
No scene-graph data is used during source-domain training.
}
\begin{tabular*}{\linewidth}{@{\extracolsep{\fill}} lcc @{}}
\toprule
Pretraining Setting & \textbf{SGG} & $\Delta$ \\
\midrule
Base (no domain training) & 19.10 & -- \\
Event-only (unified schema) & 21.47 & +2.37 \\
\bottomrule
\end{tabular*}

\label{tab:cross_domain_transfer}
\end{table}

\begin{figure}[h]
\centering
\includegraphics[width=0.95\linewidth]{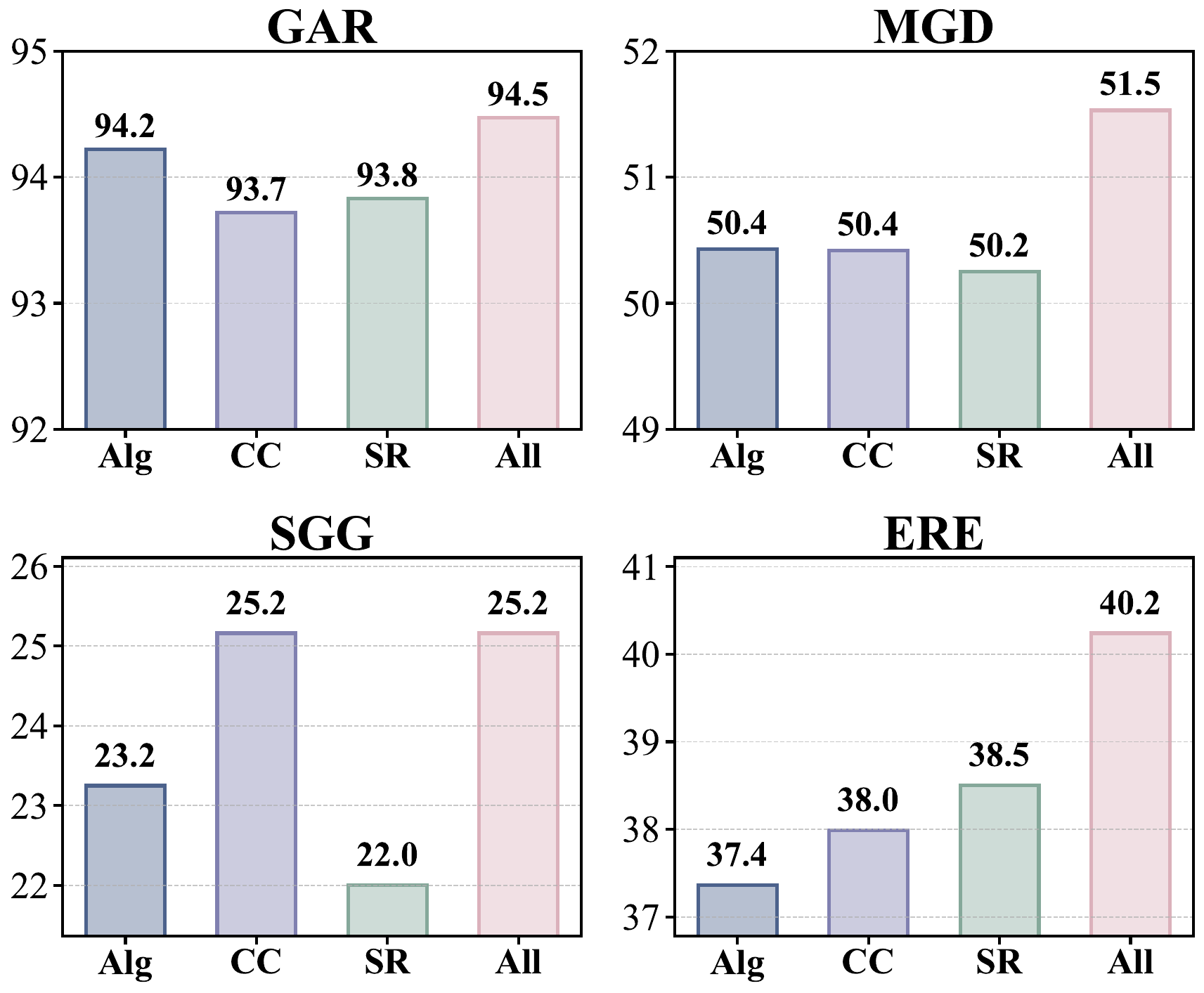}
\caption{
\textbf{Contribution of different interleaving supervision types. }
Metrics are averaged accuracy for GAR, BLEU-4 for MGD, PCIs R@50 for SGG, and
macro-averaged F1 for ERE.
}
\label{fig:cross-task}
\end{figure}

\subsubsection{Interleaving Strategy Analysis}
\label{sec:exp_interleaving}

\textbf{Cross-task Influence.} 
To analyze how different cross-role training instantiations contribute to
learning, we vary the composition of role-based exposure sources while keeping
the overall training budget fixed. Specifically, we consider three types of
role-based supervision: graph algorithmic (Alg), consistency checking (CC), and
subgraph retrieval (SR), together with their combination. All three instantiate
the \textsc{understand} side of the role function $\rho$ defined in
Section~\ref{sec:interleaved}: each takes an existing graph $G \in
\mathcal{G}_s$ as input and produces a non-graph prediction, thereby
complementing the \textsc{generate} side already covered by SGG and ERE among
the main tasks and exposing the same persistent graph to multiple functional
roles. Alg requires structural reasoning over graphs (e.g., connectivity),
encouraging preservation of global structure. CC presents the original modality
input (text or image) together with a candidate graph and predicts whether they
are consistent; negative examples are constructed by perturbing the graph,
promoting alignment between graph structure and underlying inputs. SR operates
on scene and event graphs, requiring the model to recognize structurally
meaningful subgraphs, encouraging localized structural reasoning and
compositional reuse.
Figure~\ref{fig:cross-task} shows the resulting performance changes relative to
unified multi-task training without role-based interleaving.
Gains are not uniform, but relate systematically to the domain structural
characteristics: highly constrained graph domains show smaller improvements,
whereas more weakly constrained domains benefit more from additional
cross-role structural exposure. Effects also depend on the supervision type, with
consistency checking and subgraph retrieval often yield stronger gains,
particularly when supervision is grounded in the same evidence modality.
These trends indicate that role-based interleaving reshapes
representation-level structural pressures on the persistent graph rather
than uniformly enhancing all tasks.

\begin{figure}[h]
\centering
\includegraphics[width=1.0\linewidth]{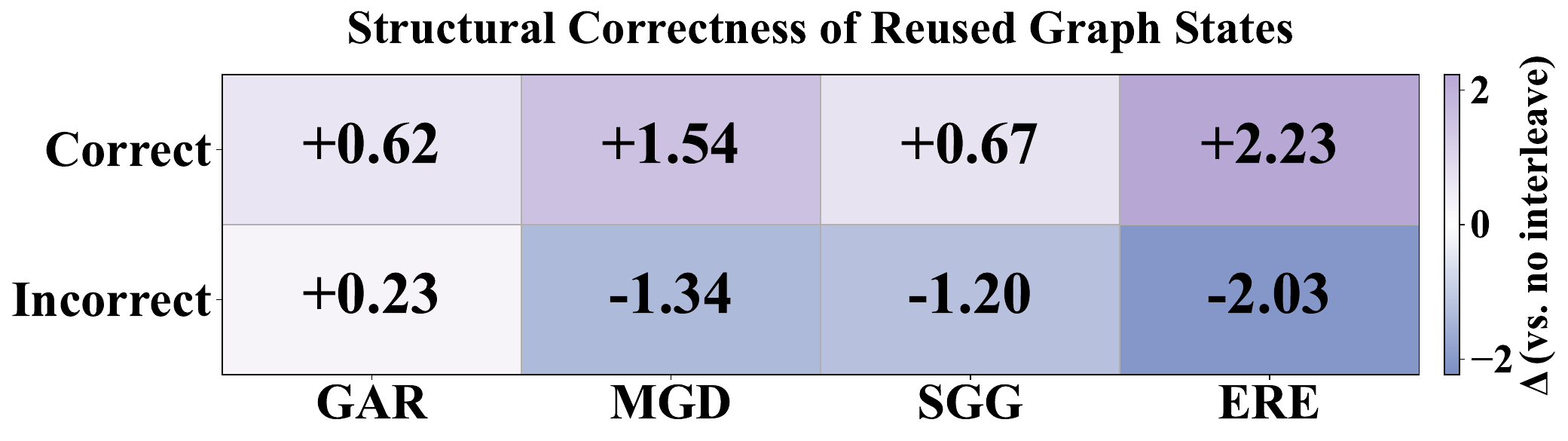}
\vspace{-0.2in}
\caption{\textbf{Effect of structural correctness of reused graph.}
Performance change ($\Delta$ vs. unified multi-task) for structurally correct and incorrect graphs. Metrics are averaged accuracy for GAR, BLEU-4 for MGD, PCIs R@50 for SGG, and
macro-averaged F1 for ERE.}
\label{fig:correctness}
\end{figure}

\textbf{Structural Correctness of Reused Graphs.} 
We test whether the gains from role-based interleaving depend on structural
coherence rather than superficial serialization.
Persistent graphs reused under multiple functional roles are replaced with
structurally incorrect variants that preserve node and edge labels but disrupt
relational connectivity.
As shown in Figure~\ref{fig:correctness}, performance gains largely disappear
when structurally incorrect graphs are used.
This contrast indicates that cross-role training is sensitive to the relational
organization of the graph, and that malformed structures introduce misleading signals at the representation level.

\begin{figure}[h]
\centering
\includegraphics[width=1.0\linewidth]{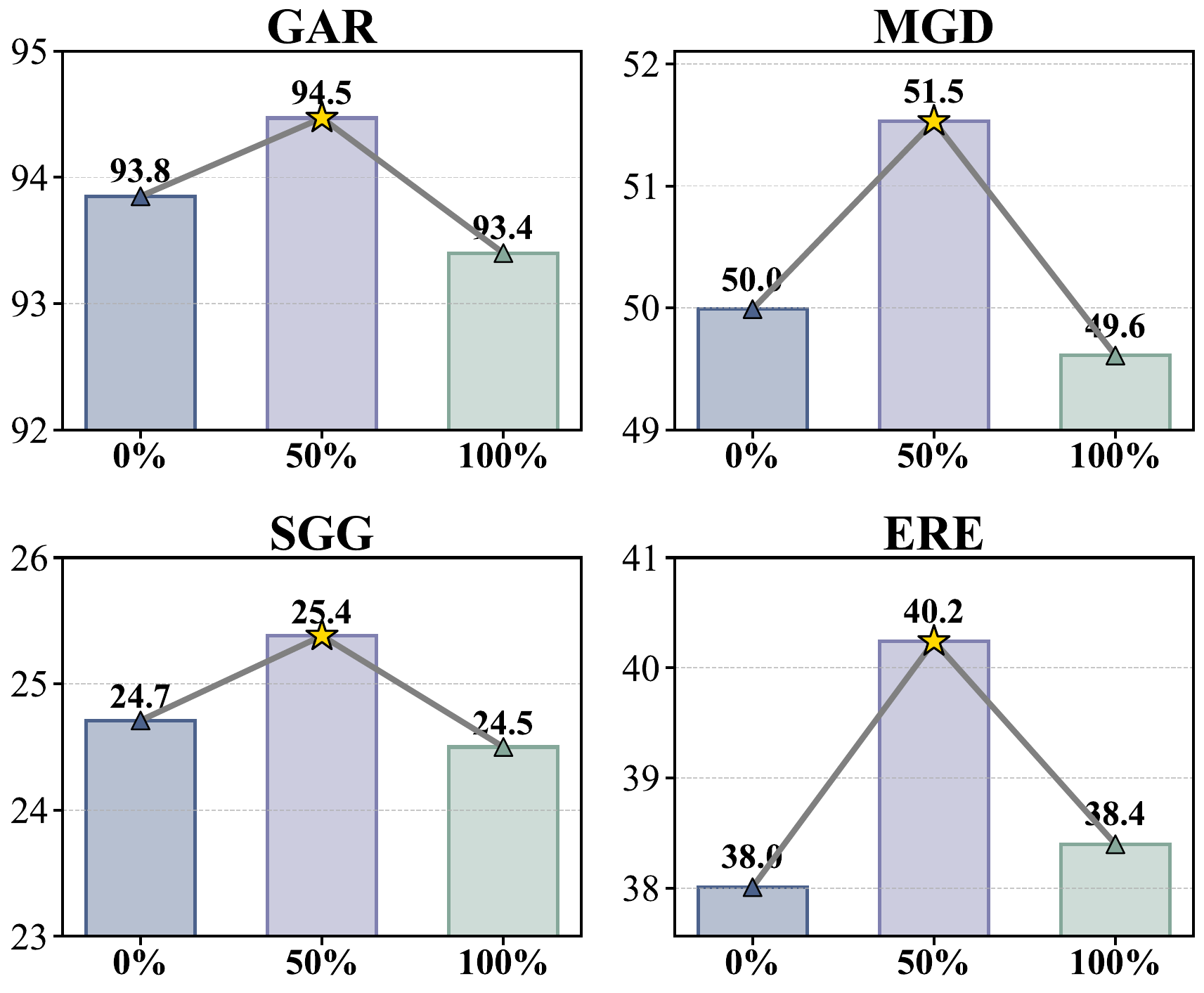}
\vspace{-0.2in}
\caption{Effect of interleaving proportion. 
Performance of each domain as the ratio of newly introduced interleaved training instances to the unified multi-task data increases (0, 50\%, 100\%). Metrics are averaged accuracy for GAR, BLEU-4 for MGD, PCIs R@50 for SGG, and averaged F1 for ERE.}
\label{fig:proportion}
\end{figure}

\textbf{Effect of Role-based Interleaving Proportion.}
Finally, we analyze how the relative proportion of role-based exposure affects
performance. As shown in Figure~\ref{fig:proportion}, we vary the ratio of
training instances in which persistent graphs are exercised under
multiple functional roles to those drawn from standard unified multi-task
training. Moderate levels of cross-role exposure consistently yield the
greatest improvements, whereas excessive role-based interleaving degrades
performance by weakening task-specific optimization signals. This trend
indicates that effective role-based training requires balancing structural
exposure of the graph across roles with sufficient task-focused learning.
Together, these results suggest that role-based interleaving operates as a
controlled mechanism for representation reuse rather than as unrestricted task
mixing.
Results with an alternative backbone show similar trends (Appendix~\ref{sec:exp_intervl}), suggesting the gains arise from representation design and training organization rather than the backbone.

\subsubsection{Robustness to Noisy Graph Extraction}
\label{sec:exp_noise}

In realistic pipelines, scene graph generators, event extractors, and parsers
all produce structurally imperfect graphs. To probe whether G-Substrate's
gains depend on clean extractions, we perturb a fixed proportion of edges in
the graphs reused during interleaved training while leaving primary task data
unperturbed. The perturbations include relation-label replacement, entity
substitution, subject--object swapping, and edge deletion. Full results
across noise levels $\{0\%, 10\%, 20\%, 30\%\}$ are in
Appendix~\ref{app:noise}. We observe graceful degradation. Under 20\%
corruption, G-Substrate still exceeds clean NMT by 50.74 to 48.11 on MGD and
by 39.74 to 38.02 on ERE, and it remains competitive on GAR; even at 30\%
noise, three of four domains stay close to or above clean NMT. SGG is the
exception, being more sensitive at all noise levels, likely because scene
graphs are structurally compact. Consistent with
Figure~\ref{fig:correctness}, complete corruption reverses gains but partial
noise does not, indicating that cross-role reuse acts as a structural
regularizer: only patterns reinforced across roles are retained, so noise in
any single context cannot dominate the learned representation.

%% file: sections/related_work.tex
\section{Related Work}
\label{sec:related_work}

\noindent\textbf{Graphs as a ubiquitous but task-bound tool.}
Graph-structured representations have become a standard modeling device across diverse domains
~\cite{GRAPH-R1:DBLP:journals/corr/abs-2508-20373,DBLP:journals/corr/abs-2509-24260,DBLP:conf/kdd/00010TL24,DBLP:conf/acl/WangWH00M24,DBLP:conf/nips/WangFHTHT23,DBLP:conf/coling/YuanLWQ25,zhang2019heterogeneous,ju2022grape}.  
Some works focus on inducing graphs from perceptual or linguistic inputs, such as scene graph generation from images and event graph extraction from text
~\cite{ESGG:DBLP:conf/eccv/ChenWLZC24,FSGG:DBLP:conf/cvpr/LiZLC024,R-SGG:DBLP:conf/aaai/LiuLW025,LLaVA-SpaceSGG:DBLP:conf/wacv/XuWZLO25,zhang2022look}. 
Other works adopt graph-conditioned reasoning paradigms, including molecular understanding and structured semantic prediction
~\cite{Mol-LLaMA:DBLP:journals/corr/abs-2502-13449,GIT-Mol:DBLP:journals/cbm/LiuRTR24,LLaMo:DBLP:conf/nips/ParkBKK24,guo2021few,guo2020graseq}.  
Despite the recurrence of similar structural patterns across domains, existing systems almost universally treat graph structure as a \emph{task-scoped artifact}: graphs are constructed to satisfy a particular objective, optimized within a single task pipeline, and discarded thereafter. Consequently, graph representations do not function as reusable state across heterogeneous learning contexts, and structural regularities are repeatedly rediscovered rather than accumulated.

\noindent\textbf{Multi-task learning.}
Multi-task learning (MTL) has long been studied as a paradigm for coordinating learning across related tasks~\cite{DBLP:journals/corr/Ruder17a,DBLP:journals/tkdd/AkhtarCE20,DBLP:journals/csur/YuanLZ25,DBLP:journals/tkde/ZhangY22,sanh2021multitask}. 
More recently, large language models and vision--language models have significantly extended this paradigm by leveraging large-scale pretraining, unified architectures, and instruction-based or prompt-based task formulations to support broad task generalization~\cite{DBLP:conf/iclr/KongF0H0CZ25,DBLP:conf/naacl/SunMFMT25,DBLP:conf/nips/WangZCZ024,DBLP:conf/nips/He0SC0LBH24,DBLP:conf/iclr/0057FKLT0Z24,wang2025generative,lu2019vilbert,tan2019lxmert,chen2020uniter}.  
Despite these advances, most existing LLM- and VLM-based multi-task frameworks rely on implicit knowledge storage in model parameters~\citep{zhang2026instruction,khashabi2020unifiedqa,mishra2022cross}, enabled by shared objectives and architectures, rather than on explicitly modeling and reusing structured representations across tasks. 
Consequently, while effective at multi-task prediction, they fall short of exploiting recurring graph structures across tasks as an explicit and reusable source of inductive bias.


\vspace{-0.15in}
\paragraph{Graph as a unified substrate across modalities.}
We argue that existing limitations stem from representation design. We therefore treat graph structure as a persistent intermediate state shared across domains and modalities, instead of a task-bound interface, enabling structural knowledge to accumulate and transfer across learning tasks. Appendix~\ref{app:related_work} provides further discussion.

%% file: sections/conclusion.tex
\section{Conclusion}

Graph structures arise across diverse domains, modalities, and tasks, but are
typically optimized in isolated learning contexts and discarded thereafter,
preventing them from serving as persistent intermediate representations. We
argue that this limitation stems from a task-centric organization of learning
that treats intermediate structure as disposable rather than reusable. To
address this issue, we introduce G-Substrate, a framework that enables
representation reuse through two complementary mechanisms: a unified structural
space that ensures cross-task compatibility, and interleaved role-based
training that exposes the same graph to multiple functional roles. Experiments
across heterogeneous settings show that the unified structural space yields
gains once multiple tasks share the representation, role-based interleaving
further amplifies these gains, and their combination yields the most consistent
improvements. Together, these findings indicate that persistent graph representations
are a key driver of structural reuse and improved performance across
diverse learning contexts.


%% file: sections/appendix.tex
\appendix
\section*{Appendix}

\section{Empirical Motivation: Recurrence of Structural Motifs Across Domains}
\label{app:empirical_motivation}

This appendix provides empirical motivation for the representation-centric
perspective in Section~\ref{sec:substrate}.
We analyze whether coarse structural
motifs recur across heterogeneous domains when relational structure is expressed
as graph states, i.e., sets of relational tuples $\mathcal{G}=\{(u,r,v)\}$.
Our goal is to verify that (i) simple local motifs (e.g., two-hop chains and hub
structures) appear consistently across domains, while (ii) global structural
scales (e.g., path lengths) can vary in a domain-dependent but interpretable
manner.

\subsection{Datasets and Graph View}
We analyze four domains used throughout the paper:
graph algorithm~\cite{GITA:DBLP:conf/nips/WeiFJZZWK024,DBLP:conf/acl/WangWH00M24},
molecular graph description~\cite{Mol-Instructions:DBLP:conf/iclr/FangL0LH0FC24},
scene graph generation~\cite{DBLP:journals/ijcv/KrishnaZGJHKCKL17},
and event relation extraction~\cite{DBLP:conf/emnlp/WangC0PWL00LLLZ22,DBLP:conf/lrec/GlavasSMK14}.
Each instance is treated as a \emph{graph state} represented by a set of relational
tuples.
For this analysis, we abstract away task semantics and focus purely on
topology-level structure; when required by the metric, directed graphs are
converted to an undirected view.

\subsection{Structural Statistics}
We report four topology-level statistics that characterize coarse structural
properties shared across domains:
(1) \textbf{AvgDeg}: the mean node degree per graph (computed on the undirected view),
capturing \emph{relational density};
(2) \textbf{ASPL}: the average shortest path length per graph (undirected),
capturing \emph{global connectivity scale};
(3) \textbf{TwoHop}: the average number of length-2 paths
($A\!\rightarrow\!B\!\rightarrow\!C$) per graph,
capturing the prevalence of \emph{two-step compositional dependencies};
(4) \textbf{Star}: the average number of hub nodes (degree $\ge 3$) per graph,
capturing \emph{hub-centric} relational organization.
All statistics are computed by first aggregating counts within each graph and
then averaging over the dataset.

\begin{table}[!h]
\centering
\small
\begin{tabular}{lrrrr}
\toprule
Domain & AvgDeg $\uparrow$ & ASPL & TwoHop $\uparrow$ & Star $\uparrow$ \\
\midrule
Graph Algorithm  & 5.8068 & 2.0787 & 632.7410 & 16.1844 \\
Molecular Graph Description   & 2.1125 & 6.1799 & 52.5278 & 11.9700 \\
Scene Graph Generation & 1.5111 & 1.4485 & 2.4198  & 0.7078 \\
Event Relation Extraction & 1.4739 & 1.3731 & 14.9518 & 0.8642 \\
\bottomrule
\end{tabular}
\caption{Topology-level structural statistics across four domains.
AvgDeg and ASPL are per-graph averages computed on undirected views.
TwoHop and Star quantify the prevalence of two-hop chains and hub nodes,
respectively.
These statistics are reported only to establish the recurrence of coarse
structural patterns across domains, rather than to compare magnitudes or
evaluate models.}
\label{tab:empirical_motivation_stats}
\end{table}

\paragraph{Cross-domain observations.}
Table~\ref{tab:empirical_motivation_stats} reveals two complementary patterns.
First, \textbf{local structural motifs recur broadly across domains}:
all four settings exhibit non-trivial two-hop dependencies and hub nodes,
indicating that compositional relational structure is not confined to any single
modality or task formulation.
Second, \textbf{global structural scale varies in an interpretable manner}.
Molecular graphs exhibit substantially larger ASPL, consistent with their
chain-like or near-tree chemical backbones.
Algorithmic graphs are denser and contain many more two-hop dependencies,
reflecting larger graph sizes and higher branching factors.
In contrast, event and scene graphs are comparatively compact
(ASPL $\approx 1.4$) with similar relational density (AvgDeg $\approx 1.5$),
suggesting that compact relational organization can arise across both textual
(event-centric) and visual (scene-centric) sources.

\subsection{Qualitative Evidence of Shared Structural Constraints Across Tasks}
\label{app:qualitative_constraints}

To further clarify how graph structure functions as a reusable substrate across
heterogeneous tasks, we present qualitative evidence showing that
\emph{structurally identical motifs not only recur across domains, but also encode
closely aligned constraint roles}.
Specifically, we pair instances from \textbf{event graphs} (text-derived) and
\textbf{scene graphs} (vision-derived) that instantiate the same coarse structural
templates, and show that these templates impose similar relational constraints
despite differences in semantics and relation inventories.

Throughout this section, motif identity is defined purely at the level of
\emph{topology} rather than label semantics.
Our goal is not to establish semantic equivalence, but to demonstrate that
shared structural forms correspond to shared constraint interpretations across
tasks.

\paragraph{Two-hop chains ($A\!\rightarrow\!B\!\rightarrow\!C$).}
We first examine two-hop chains, which represent the minimal form of compositional
relational structure.
Across both domains, all examples instantiate the same role-aligned structural
template:
\begin{center}
\small
\[
\boxed{A}\;\rightarrow\;\boxed{B}\;\rightarrow\;\boxed{C}
\]
\end{center}
\small
\emph{Constraint role:} an intermediate state $B$ composes two relations,
imposing a mediated dependency between a source $A$ and an outcome $C$.



\begin{table}[h]
\centering
\small
\setlength{\tabcolsep}{4pt}
\begin{tabular}{p{0.42\linewidth}|p{0.20\linewidth}|p{0.34\linewidth}}
\toprule
\textbf{Event graph (text)} & \textbf{Scene graph (vision)} & \textbf{Aligned constraint role} \\
\midrule
$E7:\textit{took\_place} \!\rightarrow\! E3:\textit{ridden} \!\rightarrow\! E2:\textit{remounted}$
& \textit{board} $\rightarrow$ \textit{fence} $\rightarrow$ \textit{zebra}
& Mediated dependency via an intermediate anchor \\

$E13:\textit{received} \!\rightarrow\! E3:\textit{believed} \!\rightarrow\! E31:\textit{appeared}$
& \textit{board} $\rightarrow$ \textit{counter} $\rightarrow$ \textit{bowl}
& Composition of two relations through $B$ \\

$E10:\textit{gained} \!\rightarrow\! E31:\textit{published} \!\rightarrow\! E18:\textit{avalanches}$
& \textit{engine} $\rightarrow$ \textit{train} $\rightarrow$ \textit{track}
& Two-step mediated relational path \\
\bottomrule
\end{tabular}
\caption{Cross-domain two-hop chain pairs.
Although event and scene graphs differ in semantics and relation types, both
instantiate the same compositional constraint: an intermediate node mediates
dependencies between a source and an outcome.}
\label{tab:twohop_constraint_pairs}
\end{table}

Across all pairs, the identity of the intermediate node $B$ differs in meaning
(e.g., temporal anchoring in event graphs versus spatial mediation in scene
graphs), yet its \emph{structural role} remains invariant: it serves as a
compositional bottleneck that constrains how two relations interact.
This consistency highlights that two-hop chains encode similar constraint
semantics across tasks.

\paragraph{Hub / star motifs (degree $\ge 3$).}
We next examine hub motifs, which capture cases where a single node participates
in many relations.
Across both domains, these instances instantiate the same star-like template:
\begin{center}
\small
\[
\boxed{H}\;\leftrightarrow\;\{n_i\}_{i=1}^{k},\quad k=\deg(H)
\]
\end{center}
\small
\emph{Constraint role:} a shared anchor $H$ simultaneously constrains multiple
relations, enforcing global consistency across dependent nodes.
\begin{table}[h]
\centering
\small
\setlength{\tabcolsep}{5pt}
\begin{tabularx}{\linewidth}{lcc|lcc|X}
\toprule
\multicolumn{3}{c|}{\textbf{Event graph}} &
\multicolumn{3}{c|}{\textbf{Scene graph}} &
\textbf{Aligned constraint role} \\
\cmidrule(lr){1-3}\cmidrule(lr){4-6}
Hub & Deg & Rel. type & Hub & Deg & Rel. type &  \\
\midrule
$E6:\textit{received}$ & 4 & temporal
& \textit{counter} & 3 & spatial/support
& Central anchor for multiple relations \\

$E3:\textit{ridden}$ & 6 & temporal
& \textit{guy} & 5 & part-of/attribute
& Shared reference point across dependents \\

$E31:\textit{published}$ & 20 & temporal
& \textit{train} & 6 & part-of/spatial
& Shared anchor coordinating multiple relations \\
\bottomrule
\end{tabularx}
\caption{Cross-domain hub motif pairs with aligned constraint interpretations.
Despite different relation semantics, hubs in both domains act as shared anchors
that impose multi-relation consistency.}
\label{tab:hub_constraint_pairs}
\end{table}
In both event and scene graphs, hub nodes function as structural anchors rather
than task-specific artifacts.
They concentrate relational constraints around a central node, allowing multiple
relations to be coordinated through a shared reference point.
This role remains consistent even though the surrounding relations encode
different semantics (temporal, spatial, or part-of).

Together, these paired examples demonstrate that recurring structural motifs
across tasks encode not only similar topological patterns, but also closely
aligned \emph{constraint roles}.
This observation supports the representation-centric view adopted in
Section~\ref{sec:substrate}: graph structure operates as a reusable intermediate
substrate at the level of relational organization, abstracting away from
task- or modality-specific semantics while preserving constraint-level meaning.

\section{Concrete Instantiation of the Unified Graph Representation}
\label{app:graph_representation}

This section provides a concrete instantiation of the shared graph
representation described in Section~\ref{sec:substrate}. The goal of this
instantiation is not to prescribe a canonical graph format, but to illustrate
one practical realization of the structural form used in \textbf{G-Substrate}.

In our experiments, each graph is represented as a collection of uniquely
identified entities and typed, directed relations defined over ordered pairs of
entities. Optional attributes may be attached to entities or relations when
required by specific tasks, but are treated as auxiliary annotations and do not
modify the underlying relational topology. Structural identity is therefore
determined solely by relational connectivity, which is the property that must
remain stable for graphs to be reusable across tasks and functional roles.

This structural form is used uniformly across tasks without encoding
task-specific semantics, execution procedures, or optimization objectives. The
same graph may be consumed as an intermediate representation in graph
understanding tasks or produced as an output in graph generation tasks.
Differences between tasks are expressed through prompts, supervision signals,
and evaluation protocols, rather than through modifications to the graph
structure itself.

We emphasize that this instantiation represents only one possible realization of
the shared graph representation. \textbf{G-Substrate} does not depend on any
particular schema choice, serialization format, or internal encoding, as long
as graphs conform to a consistent entity–relation structural form that enables
reuse across tasks.

\begin{table}[h]
\centering
\small
\begin{tabular}{lll}
\toprule
\textbf{Component} & \textbf{Field} & \textbf{Description} \\
\midrule
Entity & id & Unique entity identifier (e.g., E1, E2) \\
       & type & Optional entity category or label \\
\midrule
Relation & subject & Source entity identifier \\
         & predicate & Typed relation label \\
         & object & Target entity identifier \\
\midrule
Attribute & key & Optional attribute name \\
          & value & Attribute value \\
\bottomrule
\end{tabular}
\caption{Minimal structural primitives used in one realization of the shared graph representation.}
\label{tab:schema}
\end{table}

\paragraph{Graph representations across domains.}

Although instantiated in different domains, all graphs used in our experiments
conform to the same entity–relation abstraction: uniquely identified entities
connected by typed relations over ordered pairs, with optional auxiliary
attributes.

\begin{table}[h]
\centering
\small
\begin{tabular}{llll}
\toprule
\textbf{Domain} & \textbf{Entity Example} & \textbf{Relation Example} & \textbf{Structural Role} \\
\midrule
Graph Algorithmic 
& node id (e.g., 5, 17) 
& \textsc{connected} 
& Pure topology \\

Molecular Graph 
& atom (C, O, C\_aromatic) 
& \textsc{single-bond}, \textsc{aromatic-bond} 
& Chemical interaction structure \\

Event Graph 
& event mention (\textit{destroyed}, \textit{displaced}) 
& \textsc{before}, \textsc{after} 
& Temporal / causal dependency \\

Scene Graph 
& object / part (cat, box, eye) 
& \textsc{on}, \textsc{has}, \textsc{of} 
& Spatial / semantic interaction \\
\bottomrule
\end{tabular}
\caption{Examples of how different domains instantiate the same entity–relation structural abstraction. Variation lies in label vocabularies and supervision, while the relational form remains consistent.}
\label{tab:cross_domain_schema}
\end{table}

Across domains, variation lies in label vocabularies and supervision protocols
rather than in structural form. Structural identity is determined solely by
relational connectivity over entity identifiers, enabling graphs to be reused
across tasks with different functional roles without structural translation. 
While G-Substrate does not rely on any specific representation choice for its
validity, different realizations may induce different inductive biases and
therefore lead to quantitative differences in downstream performance. We
empirically study this effect in Section~\ref{sec:experiments}.

\section{Task Coverage and Framework Instantiation}
\label{app:coverage}
This section summarizes how the \textbf{G-Substrate} framework is instantiated across
different task settings.
Rather than enumerating dataset-specific configurations, we focus on how graph
representations are generated, understood, and reused across tasks under the
structural substrate defined in the main text.

\begin{tabularx}{\linewidth}{l l l X}
\toprule
\textbf{Task Setting} 
& \textbf{Input Modality} 
& \textbf{Graph Role} 
& \textbf{Reuse Pattern} \\
\midrule

Scene Graph Generation
& Image 
& Generation 
& -- \\

Event Relation Extraction
& Text 
& Generation 
& -- \\

Molecular Graph Description
& Structured Molecule Input 
& Unstanding 
& -- \\

Graph Algorithm Task
& Graph 
& Unstanding 
& Reuses graphs produced in scene graph generation and event relation extraction\\

Subgraph Retrieve Task
& Graph  
& Unstanding 
& Reuses graphs produced in scene graph generation and event relation extraction\\

Cross-Modal Consistency
& Image + Graph  
& Unstanding 
& Bidirectional reuse between perception and structure \\
\bottomrule
\end{tabularx}

Across the task instantiations considered in this work, graphs satisfy the
same structural admissibility constraints defined by the unified schema, allowing
Graph generated in one setting to remain structurally compatible with
others.
Differences between tasks arise from how graphs are generated and
understood, rather than from changes to the underlying structural constraints. For cross-modal consistency tasks, we additionally introduce controlled
structural perturbations to a subset of the reused graph to construct
negative examples. These perturbations modify relational connectivity while
preserving surface-level elements, enabling the model to distinguish
structurally coherent graphs from inconsistent ones. This design ensures that
consistent supervision provides meaningful structural learning signals rather
than relying solely on positive reuse cases.

\section{Training Paradigm Definitions}
\label{app:training_paradigms}

This section provides operational definitions of the training paradigms
summarized in Section~\ref{sec:exp_paradigms}. All paradigms use the same
backbone model, optimizer, training schedule, data mixture, and total training
budget. They differ only in (i) the representational constraints applied to graph
states and (ii) how graph states are exposed to functional roles during training.

\paragraph{Naive single-task.}
Each task is trained independently using its original task-specific graph
representation and supervision objective. Training batches contain examples
from a single task only, and graph states are constructed, optimized, and
consumed exclusively in that task. Graphs are optimized under a single
functional role, and no cross-role exposure occurs.

\paragraph{Unified single-task.}
Each task remains trained in isolation, but graphs are represented using
the unified structural schema described in Section~\ref{sec:unified_schema}.
Although the same structural admissibility constraints apply across tasks,
graphs are never exposed outside their originating task. Learning signals
remain task-specific, and graphs still operate under a single functional
role, so no reuse pressure is present.

\paragraph{Naive multi-task.}
All tasks are jointly trained by sampling batches from multiple tasks under
their native graph formats. Model parameters are updated across tasks, but graphs are still constructed and can be understood only in their originating tasks.
Graphs are therefore optimized under task-specific roles, without
structural alignment or cross-role reuse.

\paragraph{Unified multi-task (schema only).}
Tasks are jointly trained while graphs are expressed using the unified
structural schema. This imposes a common set of structural admissibility
constraints across tasks, aligning graph representations at the structural
level. However, graphs remain tied to their task of origin and are not
exposed to different functional roles. Structural compatibility is
established, but no cross-role reuse occurs.

\paragraph{Naive multi-task + interleave.}
Interleaved role-based training is introduced: the graph produced under one
task-role instantiation may be reused as inputs under another task-role
instantiation. This exposes the same graph state to multiple functional roles
during training. However, graphs retain their task-specific formats, and
no unified structural admissibility constraint is imposed. Cross-role reuse
occurs, but under heterogeneous structural conventions.

\paragraph{G-Substrate (Unified + interleave).}
Our full framework combines the unified structural schema with interleaved
role-based training. Graphs satisfy a common set of structural
admissibility constraints and are explicitly reused under multiple functional
roles during training. Learning, therefore, applies consistent pressure toward
graph representations that remain structurally compatible and reusable across
heterogeneous task contexts.

\section{Hyperparameter Configuration}
\label{app:hyperparams}

Table~\ref{tab:hyperparams} summarizes the shared hyperparameter configuration
used across all experiments, including task-isolated training, naive multi-task
learning, and the proposed \textbf{G-Substrate} framework.
All compared methods use identical model backbones, optimization settings,
training budgets, and decoding configurations.
All experiments are performed on a server with four NVIDIA A100 GPUs (40GB each).
Fine-tuning is implemented using the LLaMA-Factory framework.

\begin{table}[!h]
\centering
\small
\begin{tabular}{l c}
\toprule
\textbf{Hyperparameter} & \textbf{Value} \\
\midrule
Backbone models & Qwen3-VL-2B \\
Finetuning type & Full-parameter training \\

Vision tower frozen & Yes \\
MM projector frozen & No \\
Language model frozen & No \\

Optimizer & AdamW ($\beta_1{=}0.9$, $\beta_2{=}0.98$) \\
Learning rate & $8 \times 10^{-6}$ \\
Weight decay & 0.01 \\
Learning rate schedule & Cosine decay \\
Warmup ratio & 0.10 \\
Max input length & 2048 tokens \\
Mixed precision & bfloat16 \\
Per-device batch size & 1 \\
Gradient accumulation steps & 32 \\
Effective batch size & 64 sequences \\
Training epochs & 2 \\
Max gradient norm & 1.0 \\
Random seed & 42 \\
Decoding strategy & Greedy decoding (temperature = 0) \\
\bottomrule
\end{tabular}
\caption{Shared training configuration used across experiments unless otherwise specified.}
\label{tab:hyperparams}
\end{table}

\section{Detailed Experimental Results}
\label{app:full_results}

This appendix reports detailed results under different training paradigms of our framework, providing per-task and per-dataset breakdowns that complement the main experimental findings.

\begin{table}[!h]
\centering
\small
\begin{tabular}{lccccc}
\toprule
Method &
Connectivity &
Cycle &
Shortest Path &
Matching &
Overall \\
\midrule
Naive single-task 
& 99.44
& 92.18
& 38.27
& 92.05
& 92.89\\
Unified single-task 
& 97.80
& 94.72
& 37.14
& 85.98
& 90.43\\
Naive multi-task 
& 99.71
& 94.72
& 41.27
& 92.21
& 93.01\\
Unified multi-task 
& 98.09
& 96.19
& 45.02
& 94.23
& 93.85\\
Naive multi-task + interleave 
& 98.27
& 93.86
& 43.83
& 91.92
& 93.24\\
G-Substrate 
& 98.41
& 96.97
&48.59
&94.54
&94.47\\
\bottomrule
\end{tabular}
\caption{Detailed results on graph algorithmic tasks.}
\label{tab:ga}
\end{table}

\begin{table}[!h]
\centering
\small
\begin{tabular}{lcc}
\toprule
Method & BLEU-4 & ROUGE-L \\
\midrule
Naive single-task 
& 48.59
& 66.65\\
Unified single-task 
& 47.35
& 65.64\\
Naive multi-task 
& 48.11
&66.11\\
Unified multi-task 
&49.99
&67.36\\
Naive multi-task + interleave 
& 48.63
& 64.98\\
G-Substrate 
&51.53
&68.47\\
\bottomrule
\end{tabular}
\caption{Detailed results on molecular graph description.}
\label{tab:mgd}
\end{table}

\begin{table}[!h]
\centering
\small
\begin{tabular}{lcccc}
\toprule
Method &
PCIs R@50 &
PCIs mR@50 &
SGCLs R@50 &
SGCLs mR@50 \\
\midrule
Naive single-task 
& 23.74
& 7.78
&11.95
&4.10\\
Unified single-task 
& 22.43
& 5.84
&10.43
&4.01\\
Naive multi-task 
& 24.68
& 7.49
& 13.57
& 4.00\\
Unified multi-task 
&24.71
&7.00
&13.78
&4.80\\
Naive multi-task + interleave 
& 24.02
& 7.64
& 13.24
& 4.30\\
G-Substrate 
&25.38
&8.67
&14.07
&5.30\\
\bottomrule
\end{tabular}
\caption{Detailed results on scene graph generation.}
\label{tab:sgg}
\end{table}

\begin{table*}[!h]
\centering
\small
\setlength{\tabcolsep}{4pt}
\begin{tabular}{l ccc ccc ccc ccc}
\toprule

& \multicolumn{9}{c}{\textbf{MAVEN-ERE}} 
& \multicolumn{3}{c}{\textbf{HiEve}} \\
\cmidrule(lr){2-10}
\cmidrule(lr){11-13}

\textbf{Method}
& \multicolumn{3}{c}{Subevent}
& \multicolumn{3}{c}{Temporal}
& \multicolumn{3}{c}{Causal}
 \\

\cmidrule(lr){2-4}
\cmidrule(lr){5-7}
\cmidrule(lr){8-10}
\cmidrule(lr){11-13}

& P & R & F1
& P & R & F1
& P & R & F1
& P & R & F1 \\
\midrule

Naive single-task 
&43.44
&42.27
&39.65
&39.05
&55.22
&41.60
&31.45
&30.87
&27.70
&15.10
&30.95
&17.10\\
Unified single-task 
&52.80
&44.63
&45.45
&37.84
&42.79
&33.29
&38.07
&29.45
&30.22
&13.14
&21.83
&14.28\\
Naive multi-task
&40.44
&39.27
&36.87
&37.05
&53.22
&39.14
&37.45
&37.87
&37.02
&15.29
&30.41
&18.78\\
Unified multi-task 
&56.36
&53.52
&51.89
&44.74
&48.84
&40.05
&45.41
&43.53
&40.75
&18.34
&25.38
&19.37\\
Naive multi-task + interleave 
& 50.43
& 47.32
& 45.74
& 39.41
& 42.89
& 38.86
& 41.44
& 43.24
& 37.99
& 20.43
& 24.74
&21.36 \\
G-Substrate 
& 58.91
& 52.18
& 52.20
& 46.98
& 51.69
& 42.68
& 48.89
& 40.09
& 40.91
& 22.25
& 35.74
& 25.15\\

\bottomrule
\end{tabular}
\caption{Event relation extraction results on MAVEN-ERE and HiEve.}
\label{tab:event_results}
\end{table*}

The details from table~\ref{tab:ga} to table~\ref{tab:event_results} results across domains reveal a consistent interaction between representation format and training organization. In task-isolated settings, enforcing the unified schema alone does not yield gains and can even reduce performance, as structural constraints are not exercised beyond a single objective. Under multi-task learning, however, unified representations become beneficial, indicating that structural alignment matters once graph states are exposed to multiple learning contexts.
The full G-Substrate framework further improves over both naive and unified multi-task baselines, with the largest gains observed in tasks that rely on multi-step relational composition, such as shortest-path reasoning, rare scene-graph relations, and event substructure modeling. By contrast, naive interleaving without schema-level alignment provides only limited and unstable improvements. These patterns suggest that performance gains arise not from task mixing alone, but from organizing learning so that structurally admissible graph states are reused across heterogeneous roles.

\section{Generality Across Model Backbones}
\label{sec:exp_intervl}

To examine whether the observed performance gains are specific to a particular
vision--language model backbone, we conduct a lightweight transfer study using an alternative
vision--language model backbone from a different model family.
This analysis is intended as a robustness check rather than an exhaustive
model comparison.

We repeat a subset of the main experiments using InternVL3\_5-2B-HF under the same
training recipe, data composition, and evaluation protocol as in the main paper.
Specifically, we compare task-isolated training, naive multi-task learning,
and the full \textbf{G-Substrate} framework, along with key component ablations.

\begin{table*}[t]
\centering
\small
\setlength{\tabcolsep}{5pt}
\begin{tabular}{lcccc|cc|c|cccc}
\toprule

& \multicolumn{4}{c}{\textbf{GAR}} 
& \multicolumn{2}{c}{\textbf{MGD}} 
& \multicolumn{1}{c}{\textbf{SGG}} 
& \multicolumn{4}{c}{\textbf{ERE}} \\

\cmidrule(lr){2-5} \cmidrule(lr){6-7} \cmidrule(lr){8-8} \cmidrule(lr){9-12}

\textbf{Method}
& CT & CD & SP & BM
& BLEU-4 & ROUGE-L
& PCIs
& MA-S & MA-T & MA-C & HiE \\
\midrule

Naive single-task
& \underline{99.65} & 94.22 & 37.33 & 92.36
& 50.37 & 67.60
& 26.94
& 41.68 & 42.80 & 32.87& 16.70 \\

Unified single-task
& 97.86 & 94.29 & 36.96 & 92.36
& 49.89 & 67.44
& 28.15
& 52.65 & 37.82 & 34.80 & 10.74 \\

Naive multi-task
& \textbf{99.84} & 95.42 & 45.81& 92.67
& 51.45 & 68.28
& 26.59
& 43.40 & \underline{43.84} & 32.47 &  18.22\\

Unified multi-task
& 98.26 & \underline{95.98} & \underline{47.46} & \underline{94.39}
& \underline{51.77} & \underline{68.77}
& \underline{28.29}
& \textbf{57.04} & \textbf{45.62} & \textbf{41.57} & 18.78 \\

Naive multi-task + interleave
& 98.04 & 94.86 & 46.89 & 93.48
& 51.63 & 67.98
& 28.02
& 53.74 & 38.94 & 38.74 & \underline{21.36} \\

\textbf{G-Substrate}
& 98.28 & \textbf{98.28} & \textbf{50.65} & \textbf{94.54}
& \textbf{52.02} & \textbf{68.95}
& \textbf{29.01}
& \underline{55.23} & 43.04 & \underline{39.05} & \textbf{21.68}\\

\bottomrule
\end{tabular}

\caption{Task-level performance using an alternative model backbone (InternVL).
The same training recipe and evaluation metrics as in the main experiments are used. Best results are in \textbf{bold}; second-best are \underline{underlined}.}
\label{tab:backbone}
\end{table*}
Table~\ref{tab:backbone} shows that the overall trends observed in the main
experiments persist under a different model backbone. In task-isolated settings,
the unified schema alone does not consistently outperform native representations,
and in some cases slightly reduces performance, mirroring the behavior observed
with the primary backbone. This again indicates that structural alignment by
itself does not constitute an intrinsic performance advantage. 
Under multi-task training, however, unified representations become more effective.
Unified multi-task learning improves over naive multi-task training across most
domains, particularly in shortest-path reasoning (SP), scene graph metrics, and
event relation extraction. The full G-Substrate framework further
improves over both baselines, yielding the strongest or near-strongest results in
most settings. Notably, gains are most visible in tasks that require multi-step
relational composition, such as SP in GAR and subevent/causal relations in ERE,
which are structurally similar to the patterns seen with the original backbone. 
Naive multi-task with interleaving provides partial benefits but remains less
stable across domains, especially in ERE, where some metrics degrade relative to
unified multi-task training. This again suggests that cross-task exposure alone
is insufficient, and that consistent structural admissibility plays an important
role in enabling reliable reuse.

Overall, the consistency of these patterns across two architecturally distinct
vision--language backbones indicate that the improvements are not tied to a
specific model family. Instead, they stem from how relational structure is
represented and reused during training, supporting the generality of the
framework.








\section{Extended Related Work}
\label{app:related_work}

This appendix expands the discussion in Section~\ref{sec:related_work} and
situates our work within a broader landscape of research involving graph
structure in learning systems. We organize prior work according to a common task
formulations in which graphs arise, and analyze how graph representations are
constructed, optimized, and used. Across these paradigms, a recurring pattern
emerges: graph structure is typically introduced to satisfy the objective of an
individual task, and is rarely maintained as a persistent intermediate
representation that must remain compatible and reusable across tasks.

\subsection{Tasks over Structured Graphs with LLMs and VLMs}

A substantial body of work has studied graph-structured data using graph neural
networks and related graph representation learning methods. These approaches
encode relational structure through message passing, neighborhood aggregation,
and relation-aware propagation, and have been widely used to model typed
entities, relations, and structured dependencies in graph data
~\cite{DBLP:conf/kdd/ZhangSHSC19,DBLP:conf/iclr/VelickovicCCRLB18,DBLP:conf/iclr/KipfW17}. GNN-based methods have also been applied
across diverse structured domains, including molecular modeling,
knowledge-intensive reasoning, recommendation, and database systems
~\cite{DBLP:conf/ijcai/LiLLLZ25,DBLP:conf/esws/SchlichtkrullKB18}.

Recent LLM- and VLM-based methods broaden this line by studying tasks defined
over structured graph inputs. These include graph-theoretic reasoning and algorithmic
problems such as shortest path, connectivity, traversal, and combinatorial
queries, often realized through graph serialization, specialized prompting, or
graph-aware tokenization strategies
~\cite{liinstance,GRAPH-R1:DBLP:journals/corr/abs-2508-20373,
DBLP:journals/corr/abs-2509-24260,
DBLP:conf/kdd/00010TL24,
DBLP:conf/acl/WangWH00M24,
DBLP:conf/nips/WangFHTHT23,
DBLP:conf/coling/YuanLWQ25,
DBLP:journals/corr/abs-2410-19084,
tang2024grapharena,
10.1145/3690624.3709238}.  
More recent work extends such settings to multimodal regimes, where graphs are
derived from images or other perceptual signals and processed by VLMs
~\cite{GITA:DBLP:conf/nips/WeiFJZZWK024,
DBLP:journals/access/SartoriBB25,
DBLP:conf/icml/LiHS0W024,
DBLP:conf/acl/Zhu0CXYZ25,
zhao2025bridging,
zhao2025underappreciated}.  
Related efforts address tasks such as molecular graph description and reasoning,
where structured graphs are mapped to semantic outputs
~\cite{Mol-LLaMA:DBLP:journals/corr/abs-2502-13449,
GIT-Mol:DBLP:journals/cbm/LiuRTR24,
LLaMo:DBLP:conf/nips/ParkBKK24,
Mol-Instructions:DBLP:conf/iclr/FangL0LH0FC24,
yin2025mora,
10767279,
wu2025structure,
jin2025effective}.

Across these works, the graph is typically treated as a task-bounded input
object. Its representation is optimized only insofar as it supports the current
objective (e.g., algorithmic prediction or description generation). There is no
requirement that graph representations remain structurally compatible with other
tasks, nor that they serve as intermediate artifacts reused under different
learning roles.

\subsection{Graph Generation in Vision and Language}

Another major line of work focuses on generating graphs from perceptual or
linguistic inputs, such as scene graph generation from images
~\cite{ESGG:DBLP:conf/eccv/ChenWLZC24,
FSGG:DBLP:conf/cvpr/LiZLC024,
R-SGG:DBLP:conf/aaai/LiuLW025,
LLaVA-SpaceSGG:DBLP:conf/wacv/XuWZLO25,
DBLP:journals/corr/abs-2507-06510,
min2025vision,
hu2025spade,
DBLP:journals/pami/WangWYL25,
DBLP:journals/corr/abs-2504-00844,
chen2025data,
dutta2025open,
DBLP:journals/mms/HuZWG25,
DBLP:conf/mir/KongZ25,
DBLP:journals/corr/abs-2502-03856}
and event--event relation extraction from text
~\cite{hu-etal-2025-large,
MAQInstruct:DBLP:conf/www/XuSZZ25,
DBLP:conf/kdd/00010TL24,
DBLP:journals/corr/abs-2509-16722,
DBLP:journals/corr/abs-2508-20828,
tanev2025exploring,
li2025event,
DBLP:conf/emnlp/WangXXD24,
DBLP:conf/acl/0004W25,
DBLP:conf/emnlp/0001PZL24}.  
In these formulations, graphs serve as final prediction targets. Training
objectives optimize graph quality with respect to task-specific metrics, and the
Generated graphs are evaluated independently within each task context.

As a result, graphs are not required to persist beyond generation or to function
as reusable intermediate representations for other tasks. Structural
regularities learned during generation are not explicitly constrained to remain
compatible with graph-conditioned reasoning tasks.

\subsection{Multi-Task and Multi-Modal Learning}

Multi-task and multi-modal learning have been extensively studied as mechanisms
for coordinating learning across tasks and modalities
~\cite{DBLP:journals/corr/Ruder17a,
DBLP:journals/tkdd/AkhtarCE20,
DBLP:journals/csur/YuanLZ25,
DBLP:journals/tkde/ZhangY22}.  
Typical approaches emphasize parameter sharing
~\cite{DBLP:conf/cikm/PanFWHZ025,
DBLP:conf/kdd/LiuCGLWLLCSZJ0L24,
DBLP:conf/coling/LengX25},
task balancing
~\cite{DBLP:conf/nips/XiaFZJ024,
DBLP:journals/tkde/ZhaoSHZFL25,
DBLP:conf/emnlp/GongYLLCL24,jumulti},
curriculum design
~\cite{DBLP:journals/corr/abs-2505-14970,
DBLP:journals/tkde/ZhaoSHZFL25,
DBLP:conf/www/WangZC024},
and optimization heuristics
~\cite{DBLP:conf/nips/XiaFZJ024,
DBLP:journals/corr/abs-2504-07448,
DBLP:conf/coling/LengX25}. 
Recent work has also explored broader forms of foundation-model specialization,
including domain-specific VLM~\cite{ma2025sarvlm} and
skill-graph-based data selection for mathematical pretraining
~\cite{DBLP:conf/icml/00120C000Z25}.

These methods coordinate learning primarily at the level of parameters, losses,
data scheduling, or task curricula. When graph structure appears, it usually
serves as a task-specific input, output, or data-organization signal, and reuse
occurs implicitly through shared parameters rather than through explicit reuse
of intermediate graph states. In contrast, our framework treats graph states as
persistent intermediate representations that must remain structurally valid
across heterogeneous generation and understanding roles.

\subsection{Unified and Foundation Graph Models}

Graph foundation models aim to build general-purpose systems that transfer
across graph tasks and domains through large-scale pretraining, architectural
unification, and broad task coverage. Existing approaches can be broadly
categorized into three directions: \emph{graph-model–centric} methods that
extend graph neural architectures toward broader generality
~\cite{DBLP:conf/iclr/0057FKLT0Z24,
DBLP:journals/corr/abs-2412-16441,
DBLP:conf/nips/WangZCZ024,
DBLP:conf/www/YuZ0024,
DBLP:conf/nips/JiangQXZZ00X0W24}; \emph{language-model–centric} methods that
adapt LLMs to operate on graph-structured inputs or tasks
~\cite{DBLP:journals/corr/abs-2502-13562,
DBLP:journals/corr/abs-2410-14961,
DBLP:conf/iclr/KongF0H0CZ25,
DBLP:conf/acl/WangWH00M24,DBLP:conf/icml/WangZMCZ025}; and \emph{joint graph–language pretraining}
approaches that co-train graph and language representations within a unified
frameworks
~\cite{DBLP:conf/sigir/Tang00SSCY024,
DBLP:journals/corr/abs-2410-11370,
DBLP:conf/icml/Chen0JSW24,
DBLP:conf/www/ZhangSWFMXLYS24,
DBLP:conf/www/0011H0C24,
DBLP:conf/nips/WangZL024,
DBLP:conf/emnlp/Hu00WLLD24,
DBLP:journals/corr/abs-2505-15116,
DBLP:journals/corr/abs-2310-11829,wang2025can,thapaliya2025semantic,ma2025llm}.  
These models emphasize scale, pretraining diversity, and architectural
unification, aiming to improve transfer across graph tasks through shared
parameters and large training corpora. However, the graph structure in these
systems remains conditioned on task formulations: graph representations are
constructed and optimized with respect to individual task objectives, and are
not required to persist as intermediate artifacts beyond the originating task.

Our work explores a complementary axis of generalization. Rather than focusing
On how parameters or model architectures generalize across tasks, we study how
\emph{intermediate graph states themselves} can be organized to remain
structurally admissible and reusable under heterogeneous task roles. We
explicitly enforce structural compatibility and cross-task reuse of the graph
states, treating graphs as a reusable substrate in the learning process rather
than as task-bound artifacts. This perspective is orthogonal to scaling and
architectural unification, and addresses how structured representations persist
and function across learning contexts.

\section{Comparison with Gradient-Balancing Multi-Task Baselines}
\label{app:gradnorm}

To further situate G-Substrate relative to standard multi-task learning
algorithms that address task interference at the optimization level, we
compare against \textbf{GradNorm}~\cite{chen2018gradnorm}, a widely used
gradient-balancing method. GradNorm dynamically reweights task losses to
equalize gradient magnitudes across tasks, and we apply it on top of the naive
multi-task baseline (NMT). For fair comparison, we also evaluate G-Substrate
combined with GradNorm. We use $\alpha = 1.5$ and a weight learning rate of
$0.025$, following the original recipe.

\begin{table}[h]
\centering
\small
\caption{
\textbf{Comparison with gradient-balancing multi-task learning.}
We report averaged accuracy for GAR, BLEU-4 for MGD, PCIs R@50 for SGG, and
macro-averaged F1 for ERE.
}
\label{tab:gradnorm}
\begin{tabular}{lcccc}
\toprule
Method & GAR & MGD & SGG & ERE \\
\midrule
NMT & 93.01 & 48.11 & 25.36 & 38.02 \\
NMT + GradNorm & 93.43 & 49.64 & 25.35 & 35.09 \\
G-Substrate & 94.47 & 51.53 & 25.38 & 42.24 \\
G-Substrate + GradNorm & 94.39 & 52.49 & 26.41 & 40.24 \\
\bottomrule
\end{tabular}
\end{table}

Two observations follow. First, GradNorm improves NMT on GAR and MGD but
\emph{hurts} ERE by $-2.93$, because gradient-magnitude equalization assigns
near-zero weight to event-relation extraction once that loss converges faster.
This illustrates a known limitation of convergence-based reweighting under
heterogeneous task difficulty. Second, G-Substrate outperforms NMT+GradNorm on
\emph{all} four domains \emph{without any gradient balancing}, indicating that
the dominant bottleneck in our setting is \emph{representational}---how
relational structure is shared and reused---rather than optimization-level
loss balancing. Combining G-Substrate with GradNorm produces mixed effects
(MGD $+0.96$, SGG $+1.03$, ERE $-1.98$), suggesting that convergence-based
reweighting can interfere with the balanced cross-role exposure that
G-Substrate relies on. The two approaches address complementary but distinct
bottlenecks, and gradient balancing is not a substitute for explicit
representation reuse.

\section{Robustness to Noisy Graph Extraction}
\label{app:noise}

In practical pipelines, graph extraction is rarely perfect: scene graph
generators, event extractors, and parsers all produce structurally imperfect
graphs. To evaluate whether G-Substrate's gains depend on access to clean
graphs, we simulate imperfect extraction by injecting controlled structural
noise into the graphs reused during interleaved training. At each noise level,
a fixed proportion of edges is randomly perturbed through a mixture of
operations: relation-label replacement (40\%), entity substitution (30\%),
subject--object swapping (15\%), and edge deletion (15\%). Noise is applied
only to graphs used as interleaved cross-role training data; the primary task
data remains unperturbed.

\begin{table}[h]
\centering
\small
\caption{
\textbf{Robustness of G-Substrate under noisy graph extraction.}
Performance on the four domains as the proportion of perturbed edges increases
from 0\% (clean G-Substrate) to 30\%. NMT (clean) is provided as a reference.
We report averaged accuracy for GAR, BLEU-4 for MGD, PCIs R@50 for SGG, and
macro-averaged F1 for ERE.
}
\label{tab:noise}
\begin{tabular}{lcccc}
\toprule
Noise Level & GAR & MGD & SGG & ERE \\
\midrule
0\% (G-Substrate) & 94.47 & 51.53 & 25.38 & 42.24 \\
10\%              & 94.73 & 49.09 & 24.59 & 41.33 \\
20\%              & 92.10 & 50.74 & 23.29 & 39.74 \\
30\%              & 92.24 & 47.93 & 23.81 & 37.60 \\
\midrule
NMT (clean)       & 93.01 & 48.11 & 25.36 & 38.02 \\
\bottomrule
\end{tabular}
\end{table}

Performance degrades gradually with \emph{no catastrophic failure}.
G-Substrate under 20\% noise still outperforms clean NMT on MGD (50.74 vs.\
48.11) and ERE (39.74 vs.\ 38.02), and remains competitive on GAR (92.10 vs.\
93.01). SGG is more sensitive to noise, dropping below clean NMT at all noise
levels, likely because scene graphs are structurally compact (average 1.5
edges per relation, Table~\ref{tab:sec2_structural_signal}) and thus more
affected by per-edge perturbation. Even at 30\% noise, three of four domains
remain close to or above clean NMT levels.

This result is consistent with Figure~\ref{fig:correctness}: complete
structural corruption reverses gains, but partial noise leads to graceful
degradation, indicating that G-Substrate does not require perfect graph
extraction to remain effective. This robustness likely arises because
cross-role reuse acts as a structural regularizer: only structurally
consistent patterns shared across tasks are reinforced, so noise in any
single context does not dominate the learned representation.

%% file: example_paper.bib
@inproceedings{DBLP:conf/kdd/00010TL24,
  author       = {Nuo Chen and
                  Yuhan Li and
                  Jianheng Tang and
                  Jia Li},
  title        = {GraphWiz: An Instruction-Following Language Model for Graph Computational
                  Problems},
  booktitle    = {SIGKDD},
  year         = {2024}
}

@inproceedings{GITA:DBLP:conf/nips/WeiFJZZWK024,
  author       = {Yanbin Wei and
                  Shuai Fu and
                  Weisen Jiang and
                  Zejian Zhang and
                  Zhixiong Zeng and
                  Qi Wu and
                  James T. Kwok and
                  Yu Zhang},
  title        = {{GITA:} Graph to Visual and Textual Integration for Vision-Language
                  Graph Reasoning},
  booktitle    = {NeurIPS},
  year         = {2024}
}

@article{GRAPH-R1:DBLP:journals/corr/abs-2508-20373,
  author       = {Yuyao Wang and
                  Bowen Liu and
                  Jianheng Tang and
                  Nuo Chen and
                  Yuhan Li and
                  Qifan Zhang and
                  Jia Li},
  title        = {Graph-R1: Unleashing {LLM} Reasoning with NP-Hard Graph Problems},
  journal      = {CoRR},
  year         = {2025}
}

@article{DBLP:journals/corr/abs-2509-24260,
  author       = {Yuwei Hu and
                  Xinyi Huang and
                  Zhewei Wei and
                  Yongchao Liu and
                  Chuntao Hong},
  title        = {Rethinking and Benchmarking Large Language Models for Graph Reasoning},
  journal      = {CoRR},
  year         = {2025}
}

@article{Mol-LLaMA:DBLP:journals/corr/abs-2502-13449,
  author       = {Dongki Kim and
                  Wonbin Lee and
                  Sung Ju Hwang},
  title        = {Mol-LLaMA: Towards General Understanding of Molecules in Large Molecular
                  Language Model},
  journal      = {CoRR},
  year         = {2025}
}

@article{GIT-Mol:DBLP:journals/cbm/LiuRTR24,
  author       = {Pengfei Liu and
                  Yiming Ren and
                  Jun Tao and
                  Zhixiang Ren},
  title        = {GIT-Mol: {A} multi-modal large language model for molecular science
                  with graph, image, and text},
  journal      = {Comput. Biol. Medicine},
  year         = {2024}
}

@inproceedings{LLaMo:DBLP:conf/nips/ParkBKK24,
  author       = {Jinyoung Park and
                  Minseong Bae and
                  Dohwan Ko and
                  Hyunwoo J. Kim},
  title        = {LLaMo: Large Language Model-based Molecular Graph Assistant},
  booktitle    = {NeurIPS},
  year         = {2024}
}

@inproceedings{Mol-Instructions:DBLP:conf/iclr/FangL0LH0FC24,
  author       = {Yin Fang and
                  Xiaozhuan Liang and
                  Ningyu Zhang and
                  Kangwei Liu and
                  Rui Huang and
                  Zhuo Chen and
                  Xiaohui Fan and
                  Huajun Chen},
  title        = {Mol-Instructions: {A} Large-Scale Biomolecular Instruction Dataset
                  for Large Language Models},
  booktitle    = {ICLR},
  year         = {2024}
}

@inproceedings{R-SGG:DBLP:conf/aaai/LiuLW025,
  author       = {Tao Liu and
                  Rongjie Li and
                  Chongyu Wang and
                  Xuming He},
  title        = {Relation-aware Hierarchical Prompt for Open-vocabulary Scene Graph
                  Generation},
  booktitle    = {AAAI},
  year         = {2025}
}

@inproceedings{LLaVA-SpaceSGG:DBLP:conf/wacv/XuWZLO25,
  author       = {Mingjie Xu and
                  Mengyang Wu and
                  Yuzhi Zhao and
                  Jason Chun Lok Li and
                  Weifeng Ou},
  title        = {LLaVA-SpaceSGG: Visual Instruct Tuning for Open-Vocabulary Scene Graph
                  Generation with Enhanced Spatial Relations},
  booktitle    = {WACV},
  year         = {2025}
}

@inproceedings{ESGG:DBLP:conf/eccv/ChenWLZC24,
  author       = {Zuyao Chen and
                  Jinlin Wu and
                  Zhen Lei and
                  Zhaoxiang Zhang and
                  Chang Wen Chen},
  title        = {Expanding Scene Graph Boundaries: Fully Open-Vocabulary Scene Graph
                  Generation via Visual-Concept Alignment and Retention},
  booktitle    = {ECCV},
  year         = {2024}
}

@inproceedings{FSGG:DBLP:conf/cvpr/LiZLC024,
  author       = {Rongjie Li and
                  Songyang Zhang and
                  Dahua Lin and
                  Kai Chen and
                  Xuming He},
  title        = {From Pixels to Graphs: Open-Vocabulary Scene Graph Generation with
                  Vision-Language Models},
  booktitle    = {CVPR},
  year         = {2024}
}

@inproceedings{hu-etal-2025-large,
    title = "Large Language Model-Based Event Relation Extraction with Rationales",
    author = "Hu, Zhilei  and
      Li, Zixuan  and
      Jin, Xiaolong  and
      Bai, Long  and
      Guo, Jiafeng  and
      Cheng, Xueqi",
    booktitle = "COLING",
    year = "2025",
}

@inproceedings{MAQInstruct:DBLP:conf/www/XuSZZ25,
  author       = {Jun Xu and
                  Mengshu Sun and
                  Zhiqiang Zhang and
                  Jun Zhou},
  title        = {MAQInstruct: Instruction-based Unified Event Relation Extraction},
  booktitle    = {WWW},
  year         = {2025}
}

@article{GraphOmni:DBLP:journals/corr/abs-2504-12764,
  author       = {Hao Xu and
                  Xiangru Jian and
                  Xinjian Zhao and
                  Wei Pang and
                  Chao Zhang and
                  Suyuchen Wang and
                  Qixin Zhang and
                  Joao Monteiro and
                  Qiuzhuang Sun and
                  Tianshu Yu},
  title        = {GraphOmni: {A} Comprehensive and Extendable Benchmark Framework for
                  Large Language Models on Graph-theoretic Tasks},
  journal      = {CoRR},
  year         = {2025}
}

@article{chai2025graphllm,
  title={Graphllm: Boosting graph reasoning ability of large language model},
  author={Chai, Ziwei and Zhang, Tianjie and Wu, Liang and Han, Kaiqiang and Hu, Xiaohai and Huang, Xuanwen and Yang, Yang},
  journal={TBD},
  year={2025},
  publisher={IEEE}
}

@inproceedings{wu2025universal,
  title={Universal scene graph generation},
  author={Wu, Shengqiong and Fei, Hao and Chua, Tat-Seng},
  booktitle={CVPR},
  year={2025}
}

@inproceedings{tao2025comprehensive,
  title={A comprehensive evaluation on event reasoning of large language models},
  author={Tao, Zhengwei and Jin, Zhi and Zhang, Yifan and Chen, Xiancai and Zhao, Haiyan and Li, Jia and Liang, Bin and Tao, Chongyang and Liu, Qun and Wong, Kam-Fai},
  booktitle={AAAI},
  year={2025}
}

@inproceedings{DBLP:conf/acl/WangWH00M24,
  author       = {Jianing Wang and
                  Junda Wu and
                  Yupeng Hou and
                  Yao Liu and
                  Ming Gao and
                  Julian J. McAuley},
  title        = {InstructGraph: Boosting Large Language Models via Graph-centric Instruction
                  Tuning and Preference Alignment},
  booktitle    = {ACL},
  year         = {2024}
}

@inproceedings{DBLP:conf/nips/WangFHTHT23,
  author       = {Heng Wang and
                  Shangbin Feng and
                  Tianxing He and
                  Zhaoxuan Tan and
                  Xiaochuang Han and
                  Yulia Tsvetkov},
  title        = {Can Language Models Solve Graph Problems in Natural Language?},
  booktitle    = {NeurIPS},
  year         = {2023}
}

@inproceedings{DBLP:conf/coling/YuanLWQ25,
  author       = {Zike Yuan and
                  Ming Liu and
                  Hui Wang and
                  Bing Qin},
  title        = {GraCoRe: Benchmarking Graph Comprehension and Complex Reasoning in
                  Large Language Models},
  booktitle    = {COLING},
  year         = {2025}
}

@article{DBLP:journals/access/SartoriBB25,
  author       = {Camilo Chac{\'{o}}n Sartori and
                  Christian Blum and
                  Filippo Bistaffa},
  title        = {VisGraphVar: {A} benchmark generator for Assessing Variability in
                  Graph Analysis Using Large Vision-Language Models},
  journal      = {Access},
  year         = {2025}
}

@inproceedings{DBLP:conf/icml/LiHS0W024,
  author       = {Yunxin Li and
                  Baotian Hu and
                  Haoyuan Shi and
                  Wei Wang and
                  Longyue Wang and
                  Min Zhang},
  title        = {VisionGraph: Leveraging Large Multimodal Models for Graph Theory Problems
                  in Visual Context},
  booktitle    = {ICML},
  year         = {2024}
}

@inproceedings{DBLP:conf/acl/Zhu0CXYZ25,
  author       = {Yingjie Zhu and
                  Xuefeng Bai and
                  Kehai Chen and
                  Yang Xiang and
                  Jun Yu and
                  Min Zhang},
  title        = {Benchmarking and Improving Large Vision-Language Models for Fundamental
                  Visual Graph Understanding and Reasoning},
  booktitle    = {ACL},
  year         = {2025}
}

@article{DBLP:journals/corr/abs-2410-19084,
  author       = {Qifan Zhang and
                  Xiaobin Hong and
                  Jianheng Tang and
                  Nuo Chen and
                  Yuhan Li and
                  Wenzhong Li and
                  Jing Tang and
                  Jia Li},
  title        = {GCoder: Improving Large Language Model for Generalized Graph Problem
                  Solving},
  journal      = {CoRR},
  year         = {2024}
}

@inproceedings{DBLP:conf/icml/WangZMCZ025a,
  author       = {Zehong Wang and
                  Zheyuan Zhang and
                  Tianyi Ma and
                  Nitesh V. Chawla and
                  Chuxu Zhang and
                  Yanfang Ye},
  title        = {Beyond Message Passing: Neural Graph Pattern Machine},
  booktitle    = {ICML},
  year         = {2025}
}

@article{tang2024grapharena,
  title={Grapharena: Evaluating and exploring large language models on graph computation},
  author={Tang, Jianheng and Zhang, Qifan and Li, Yuhan and Chen, Nuo and Li, Jia},
  journal={CoRR},
  year={2024}
}

@inproceedings{10.1145/3690624.3709238,
author = {Wang, Rongzheng and Liang, Shuang and Chen, Qizhi and Zhang, Jiasheng and Qin, Ke},
title = {GraphTool-Instruction: Revolutionizing Graph Reasoning in LLMs through Decomposed Subtask Instruction},
year = {2025},
booktitle = {SIGKDD}
}

@article{zhao2025bridging,
  title={Bridging visualization and optimization: Multimodal large language models on graph-structured combinatorial optimization},
  author={Zhao, Jie and Cheong, Kang Hao and Pedrycz, Witold},
  journal={CoRR},
  year={2025}
}

@article{zhao2025underappreciated,
  title={The Underappreciated Power of Vision Models for Graph Structural Understanding},
  author={Zhao, Xinjian and Pang, Wei and Xue, Zhongkai and Jian, Xiangru and Zhang, Lei and Xu, Yaoyao and Song, Xiaozhuang and Wu, Shu and Yu, Tianshu},
  journal={CoRR},
  year={2025}
}

@article{yin2025mora,
  title={MoRA: On-the-fly Molecule-aware Low-Rank Adaptation Framework for LLM-based Multi-Modal Molecular Assistant},
  author={Yin, Tao and Zhang, Xiaohong and Zhang, Jiacheng and Huang, Li and Zhang, Zhibin and Zeng, Yuansong and Xie, Jin and Yan, Meng},
  journal={CoRR},
  year={2025}
}

@ARTICLE{10767279,
  author={Luo, Yizhen and Zhang, Jiahuan and Fan, Siqi and Yang, Kai and Hong, Massimo and Wu, Yushuai and Qiao, Mu and Nie, Zaiqing},
  journal={IEEE JBHI}, 
  title={BioMedGPT: An Open Multimodal Large Language Model for BioMedicine}, 
  year={2024},
}

@inproceedings{wu2025structure,
  title={Structure-enhanced protein instruction tuning: Towards general-purpose protein understanding with llms},
  author={Wu, Wei and Wang, Chao and Chen, Liyi and Yin, Mingze and Zhu, Yiheng and Fu, Kun and Ye, Jieping and Xiong, Hui and Wang, Zheng},
  booktitle={SIGKDD},
  year={2025}
}

@article{jin2025effective,
  title={Effective and Explainable Molecular Property Prediction by Chain-of-Thought Enabled Large Language Models and Multi-Modal Molecular Information Fusion},
  author={Jin, Chang and Guo, Siyuan and Zhou, Shuigeng and Guan, Jihong},
  journal={Journal of Chemical Information and Modeling},
  year={2025}
}

@article{DBLP:journals/corr/abs-2507-06510,
  author       = {Yupeng Hu and
                  Changxing Ding and
                  Chang Sun and
                  Shaoli Huang and
                  Xiangmin Xu},
  title        = {Bilateral Collaboration with Large Vision-Language Models for Open
                  Vocabulary Human-Object Interaction Detection},
  journal      = {CVPR},
  year         = {2025},
}

@inproceedings{min2025vision,
	title={Vision-Language Interactive Relation Mining for Open-Vocabulary Scene Graph Generation},
	author={Min, Yukuan and Yang, Muli and Zhang, Jinhao and Wang, Yuxuan and Wu, Aming and Deng, Cheng},
	booktitle={ICCV},
	year={2025}
}

@inproceedings{hu2025spade,
  title={SPADE: spatial-aware denoising network for open-vocabulary panoptic scene graph generation with long-and local-range context reasoning},
  author={Hu, Xin and Qin, Ke and Duan, Guiduo and Li, Ming and Li, Yuan-Fang and He, Tao},
  booktitle={ICCV},
  year={2025}
}

@article{DBLP:journals/pami/WangWYL25,
  author       = {Yongqi Wang and
                  Xinxiao Wu and
                  Shuo Yang and
                  Jiebo Luo},
  title        = {End-to-End Open-Vocabulary Video Visual Relationship Detection Using
                  Multi-Modal Prompting},
  journal      = {TPAMI},
  year         = {2025}
}

@article{DBLP:journals/corr/abs-2504-00844,
  author       = {Abdelrahman Elskhawy and
                  Mengze Li and
                  Nassir Navab and
                  Benjamin Busam},
  title        = {{PRISM-0:} {A} Predicate-Rich Scene Graph Generation Framework for
                  Zero-Shot Open-Vocabulary Tasks},
  journal      = {CoRR},
  year         = {2025}
}

@article{chen2025data,
  title={From Data to Modeling: Fully Open-vocabulary Scene Graph Generation},
  author={Chen, Zuyao and Wu, Jinlin and Lei, Zhen and Chen, Chang Wen},
  journal={CoRR},
  year={2025}
}

@article{dutta2025open,
  title={Open World Scene Graph Generation using Vision Language Models},
  author={Dutta, Amartya and Mehrab, Kazi Sajeed and Sawhney, Medha and Neog, Abhilash and Khurana, Mridul and Fatemi, Sepideh and Pradhan, Aanish and Maruf, M and Lourentzou, Ismini and Daw, Arka and others},
  journal={CoRR},
  year={2025}
}

@article{DBLP:journals/mms/HuZWG25,
  author       = {Yufan Hu and
                  Fang Zhang and
                  Ran Wei and
                  Junling Gao},
  title        = {Learning semantic-unified cross-modal representations for open-vocabulary
                  video scene graph generation},
  journal      = {Multim. Syst.},
  year         = {2025}
}

@inproceedings{DBLP:conf/mir/KongZ25,
  author       = {Zihan Kong and
                  Haiwei Zhang},
  editor       = {Zhongfei (Mark) Zhang and
                  Elisa Ricci and
                  Yan Yan and
                  Liqiang Nie and
                  Vincent Oria and
                  Lamberto Ballan},
  title        = {OpenSGen: Fine-Grained Relation-Aware Prompt for Open-Vocabulary Scene
                  Graph Generation},
  booktitle    = {ICMR},
  year         = {2025}
}

@article{DBLP:journals/corr/abs-2502-03856,
  author       = {Lin Li and
                  Chuhan Zhang and
                  Dong Zhang and
                  Chong Sun and
                  Chen Li and
                  Long Chen},
  title        = {Taking {A} Closer Look at Interacting Objects: Interaction-Aware Open
                  Vocabulary Scene Graph Generation},
  journal      = {CoRR},
  year         = {2025}
}

@article{DBLP:journals/corr/abs-2509-16722,
  author       = {Xiaohan Ding and
                  Kaike Ping and
                  Buse {\c{C}}arik and
                  Eugenia Ha Rim Rho},
  title        = {A Multi-Level Benchmark for Causal Language Understanding in Social
                  Media Discourse},
  journal      = {CoRR},
  year         = {2025}
}

@article{DBLP:journals/corr/abs-2508-20828,
  author       = {Jie Zhao and
                  Wanting Ning and
                  Yuxiao Fei and
                  Yubo Feng and
                  Lishuang Li},
  title        = {{GDLLM:} {A} Global Distance-aware Modeling Approach Based on Large
                  Language Models for Event Temporal Relation Extraction},
  journal      = {CoRR},
  year         = {2025}
}

@inproceedings{tanev2025exploring,
  title={Exploring the Performance of Large Language Models for Event Detection and Extraction in the Health Domain},
  author={Tanev, Hristo and Stefanovitch, Nicolas and Harmatha, Tom{\'a}{\v{s}} and Sousa, Diana F},
  booktitle={RANLP},
  year={2025}
}

@article{li2025event,
  title={Event Extraction in Large Language Model},
  author={Li, Bobo and Han, Xudong and Liu, Jiang and Ding, Yuzhe and Jing, Liqiang and Zhang, Zhaoqi and Li, Jinheng and Du, Xinya and Li, Fei and Zhang, Meishan and others},
  journal={CoRR},
  year={2025}
}

@inproceedings{DBLP:conf/emnlp/WangXXD24,
  author       = {Zimu Wang and
                  Lei Xia and
                  Wei Xjtlu and
                  Xinya Du},
  title        = {Document-level Causal Relation Extraction with Knowledge-guided Binary
                  Question Answering},
  booktitle    = {EMNLP},
  year         = {2024}
}

@inproceedings{DBLP:conf/acl/0004W25,
  author       = {Ziyang Liu and
                  Chaokun Wang},
  title        = {TeRDy: Temporal Relation Dynamics through Frequency Decomposition
                  for Temporal Knowledge Graph Completion},
  booktitle    = {ACL},
  year         = {2025}
}

@inproceedings{DBLP:conf/emnlp/0001PZL24,
  author       = {Meiqi Chen and
                  Bo Peng and
                  Yan Zhang and
                  Chaochao Lu},
  title        = {{CELLO:} Causal Evaluation of Large Vision-Language Models},
  booktitle    = {ENMLP},
  year         = {2024}
}

@article{DBLP:journals/tkdd/AkhtarCE20,
  author       = {Md. Shad Akhtar and
                  Dushyant Singh Chauhan and
                  Asif Ekbal},
  title        = {A Deep Multi-task Contextual Attention Framework for Multi-modal Affect
                  Analysis},
  journal      = {TKDD},
  year         = {2020},
}

@article{DBLP:journals/csur/YuanLZ25,
  author       = {Yuan Yuan and
                  Zhaojian Li and
                  Bin Zhao},
  title        = {A Survey of Multimodal Learning: Methods, Applications, and Future},
  journal      = {{ACM} Comput. Surv.},
  year         = {2025}
}

@article{DBLP:journals/tkde/ZhangY22,
  author       = {Yu Zhang and
                  Qiang Yang},
  title        = {A Survey on Multi-Task Learning},
  journal      = {TKDE},
  year         = {2022}
}

@inproceedings{DBLP:conf/kdd/LiuCGLWLLCSZJ0L24,
  author       = {Bingchang Liu and
                  Chaoyu Chen and
                  Zi Gong and
                  Cong Liao and
                  Huan Wang and
                  Zhichao Lei and
                  Ming Liang and
                  Dajun Chen and
                  Min Shen and
                  Hailian Zhou and
                  Wei Jiang and
                  Hang Yu and
                  Jianguo Li},
  title        = {MFTCoder: Boosting Code LLMs with Multitask Fine-Tuning},
  booktitle    = {SIGKDD},
  year         = {2024}
}

@inproceedings{DBLP:conf/cikm/PanFWHZ025,
  author       = {Dayan Pan and
                  Zhaoyang Fu and
                  Jingyuan Wang and
                  Xiao Han and
                  Yue Zhu and
                  Xiangyu Zhao},
  title        = {Contextual Attention Modulation: Towards Efficient Multi-Task Adaptation
                  in Large Language Models},
  booktitle    = {CIKM},
  year         = {2025}
}

@inproceedings{DBLP:conf/coling/LengX25,
  author       = {Yongqi Leng and
                  Deyi Xiong},
  title        = {Towards Understanding Multi-Task Learning (Generalization) of LLMs
                  via Detecting and Exploring Task-Specific Neurons},
  booktitle    = {COLING},
  year         = {2025}
}

@inproceedings{DBLP:conf/nips/XiaFZJ024,
  author       = {Yifei Xia and
                  Fangcheng Fu and
                  Wentao Zhang and
                  Jiawei Jiang and
                  Bin Cui},
  title        = {Efficient Multi-task {LLM} Quantization and Serving for Multiple LoRA
                  Adapters},
  booktitle    = {NeurIPS},
  year         = {2024}
}

@article{DBLP:journals/tkde/ZhaoSHZFL25,
  author       = {Chuang Zhao and
                  Xing Su and
                  Ming He and
                  Hongke Zhao and
                  Jianping Fan and
                  Xiaomeng Li},
  title        = {Collaborative Knowledge Fusion: {A} Novel Method for Multi-Task Recommender
                  Systems via LLMs},
  journal      = {TKDE},
  year         = {2025}
}

@inproceedings{DBLP:conf/emnlp/GongYLLCL24,
  author       = {Zi Gong and
                  Hang Yu and
                  Cong Liao and
                  Bingchang Liu and
                  Chaoyu Chen and
                  Jianguo Li},
  title        = {CoBa: Convergence Balancer for Multitask Finetuning of Large Language
                  Models},
  booktitle    = {EMNLP},
  year         = {2024}
}

@article{DBLP:journals/corr/abs-2505-14970,
  author       = {Xiaoyin Chen and
                  Jiarui Lu and
                  Minsu Kim and
                  Dinghuai Zhang and
                  Jian Tang and
                  Alexandre Pich{\'{e}} and
                  Nicolas Gontier and
                  Yoshua Bengio and
                  Ehsan Kamalloo},
  title        = {Self-Evolving Curriculum for {LLM} Reasoning},
  journal      = {CoRR},
  year         = {2025}
}

@inproceedings{DBLP:conf/www/WangZC024,
  author       = {Xin Wang and
                  Yuwei Zhou and
                  Hong Chen and
                  Wenwu Zhu},
  title        = {Curriculum Learning: Theories, Approaches, Applications, Tools, and
                  Future Directions in the Era of Large Language Models},
  booktitle    = {WWW},
  year         = {2024}
}

@article{DBLP:journals/corr/abs-2504-07448,
  author       = {Juzheng Zhang and
                  Jiacheng You and
                  Ashwinee Panda and
                  Tom Goldstein},
  title        = {LoRI: Reducing Cross-Task Interference in Multi-Task Low-Rank Adaptation},
  journal      = {CoRR},
  year         = {2025}
}

@inproceedings{DBLP:conf/iclr/0057FKLT0Z24,
  author       = {Hao Liu and
                  Jiarui Feng and
                  Lecheng Kong and
                  Ningyue Liang and
                  Dacheng Tao and
                  Yixin Chen and
                  Muhan Zhang},
  title        = {One For All: Towards Training One Graph Model For All Classification
                  Tasks},
  booktitle    = {ICLR},
  year         = {2024}
}

@article{DBLP:journals/corr/abs-2412-16441,
  author       = {Zehong Wang and
                  Zheyuan Zhang and
                  Tianyi Ma and
                  Nitesh V. Chawla and
                  Chuxu Zhang and
                  Yanfang Ye},
  title        = {Learning Cross-Task Generalities Across Graphs via Task-trees},
  journal      = {CoRR},
  year         = {2024},
}

@inproceedings{DBLP:conf/nips/WangZCZ024,
  author       = {Zehong Wang and
                  Zheyuan Zhang and
                  Nitesh V. Chawla and
                  Chuxu Zhang and
                  Yanfang Ye},
  title        = {{GFT:} Graph Foundation Model with Transferable Tree Vocabulary},
  booktitle    = {NeurIPS},
  year         = {2024}
}

@inproceedings{DBLP:conf/www/YuZ0024,
  author       = {Xingtong Yu and
                  Chang Zhou and
                  Yuan Fang and
                  Xinming Zhang},
  title        = {MultiGPrompt for Multi-Task Pre-Training and Prompting on Graphs},
  booktitle    = {WWW},
  year         = {2024}
}

@inproceedings{DBLP:conf/nips/JiangQXZZ00X0W24,
  author       = {Xinke Jiang and
                  Rihong Qiu and
                  Yongxin Xu and
                  Wentao Zhang and
                  Yichen Zhu and
                  Ruizhe Zhang and
                  Yuchen Fang and
                  Chu Xu and
                  Junfeng Zhao and
                  Yasha Wang},
  title        = {RAGraph: {A} General Retrieval-Augmented Graph Learning Framework},
  booktitle    = {NeurIPS},
  year         = {2024}
}

@inproceedings{DBLP:conf/lrec/GlavasSMK14,
  author       = {Goran Glavas and
                  Jan Snajder and
                  Marie{-}Francine Moens and
                  Parisa Kordjamshidi},
  title        = {HiEve: {A} Corpus for Extracting Event Hierarchies from News Stories},
  booktitle    = {LREC},
  year         = {2014}
}

@inproceedings{DBLP:conf/naacl/SunMFMT25,
  author       = {Yuanfu Sun and
                  Zhengnan Ma and
                  Yi Fang and
                  Jing Ma and
                  Qiaoyu Tan},
  title        = {GraphICL: Unlocking Graph Learning Potential in LLMs through Structured
                  Prompt Design},
  booktitle    = {NAACL},
  year         = {2025}
}

@article{DBLP:journals/corr/abs-2502-13562,
  author       = {Jintang Li and
                  Ruofan Wu and
                  Yuchang Zhu and
                  Huizhe Zhang and
                  Liang Chen and
                  Zibin Zheng},
  title        = {Are Large Language Models In-Context Graph Learners?},
  journal      = {CoRR},
  year         = {2025}
}

@article{DBLP:journals/corr/abs-2410-14961,
  author       = {Tianqianjin Lin and
                  Pengwei Yan and
                  Kaisong Song and
                  Zhuoren Jiang and
                  Yangyang Kang and
                  Jun Lin and
                  Weikang Yuan and
                  Junjie Cao and
                  Changlong Sun and
                  Xiaozhong Liu},
  title        = {LangGFM: {A} Large Language Model Alone Can be a Powerful Graph Foundation
                  Model},
  journal      = {CoRR},
  year         = {2024}
}

@inproceedings{DBLP:conf/iclr/KongF0H0CZ25,
  author       = {Lecheng Kong and
                  Jiarui Feng and
                  Hao Liu and
                  Chengsong Huang and
                  Jiaxin Huang and
                  Yixin Chen and
                  Muhan Zhang},
  title        = {{GOFA:} {A} Generative One-For-All Model for Joint Graph Language
                  Modeling},
  booktitle    = {ICLR},
  year         = {2025}
}

@inproceedings{DBLP:conf/sigir/Tang00SSCY024,
  author       = {Jiabin Tang and
                  Yuhao Yang and
                  Wei Wei and
                  Lei Shi and
                  Lixin Su and
                  Suqi Cheng and
                  Dawei Yin and
                  Chao Huang},
  title        = {GraphGPT: Graph Instruction Tuning for Large Language Models},
  booktitle    = {SIGIR},
  year         = {2024}
}

@article{DBLP:journals/corr/abs-2410-11370,
  author       = {Haitong Luo and
                  Xuying Meng and
                  Suhang Wang and
                  Tianxiang Zhao and
                  Fali Wang and
                  Hanyun Cao and
                  Yujun Zhang},
  title        = {Enhance Graph Alignment for Large Language Models},
  journal      = {CoRR},
  year         = {2024}
}

@inproceedings{DBLP:conf/icml/Chen0JSW24,
  author       = {Runjin Chen and
                  Tong Zhao and
                  Ajay Kumar Jaiswal and
                  Neil Shah and
                  Zhangyang Wang},
  title        = {LLaGA: Large Language and Graph Assistant},
  booktitle    = {ICML},
  year         = {2024}
}

@inproceedings{DBLP:conf/www/ZhangSWFMXLYS24,
  author       = {Mengmei Zhang and
                  Mingwei Sun and
                  Peng Wang and
                  Shen Fan and
                  Yanhu Mo and
                  Xiaoxiao Xu and
                  Hong Liu and
                  Cheng Yang and
                  Chuan Shi},
  title        = {GraphTranslator: Aligning Graph Model to Large Language Model for
                  Open-ended Tasks},
  booktitle    = {WWW},
  year         = {2024}
}

@inproceedings{DBLP:conf/www/0011H0C24,
  author       = {Zheyuan Liu and
                  Xiaoxin He and
                  Yijun Tian and
                  Nitesh V. Chawla},
  title        = {Can we Soft Prompt LLMs for Graph Learning Tasks?},
  booktitle    = {WWW},
  year         = {2024}
}

@inproceedings{DBLP:conf/nips/WangZL024,
  author       = {Duo Wang and
                  Yuan Zuo and
                  Fengzhi Li and
                  Junjie Wu},
  title        = {LLMs as Zero-shot Graph Learners: Alignment of {GNN} Representations
                  with {LLM} Token Embeddings},
  booktitle    = {NeurIPS},
  year         = {2024}
}

@inproceedings{DBLP:conf/emnlp/Hu00WLLD24,
  author       = {Zhengyu Hu and
                  Yichuan Li and
                  Zhengyu Chen and
                  Jingang Wang and
                  Han Liu and
                  Kyumin Lee and
                  Kaize Ding},
  title        = {Let's Ask {GNN:} Empowering Large Language Model for Graph In-Context
                  Learning},
  booktitle    = {EMNLP},
  year         = {2024}
}

@article{DBLP:journals/corr/abs-2505-15116,
  author       = {Zehong Wang and
                  Zheyuan Liu and
                  Tianyi Ma and
                  Jiazheng Li and
                  Zheyuan Zhang and
                  Xingbo Fu and
                  Yiyang Li and
                  Zhengqing Yuan and
                  Wei Song and
                  Yijun Ma and
                  Qingkai Zeng and
                  Xiusi Chen and
                  Jianan Zhao and
                  Jundong Li and
                  Meng Jiang and
                  Pietro Lio and
                  Nitesh V. Chawla and
                  Chuxu Zhang and
                  Yanfang Ye},
  title        = {Graph Foundation Models: {A} Comprehensive Survey},
  journal      = {CoRR},
  year         = {2025}
}

@article{DBLP:journals/corr/abs-2310-11829,
  author       = {Jiawei Liu and
                  Cheng Yang and
                  Zhiyuan Lu and
                  Junze Chen and
                  Yibo Li and
                  Mengmei Zhang and
                  Ting Bai and
                  Yuan Fang and
                  Lichao Sun and
                  Philip S. Yu and
                  Chuan Shi},
  title        = {Towards Graph Foundation Models: {A} Survey and Beyond},
  journal      = {CoRR},
  year         = {2023}
}

@inproceedings{DBLP:conf/nips/He0SC0LBH24,
  author       = {Xiaoxin He and
                  Yijun Tian and
                  Yifei Sun and
                  Nitesh V. Chawla and
                  Thomas Laurent and
                  Yann LeCun and
                  Xavier Bresson and
                  Bryan Hooi},
  title        = {G-Retriever: Retrieval-Augmented Generation for Textual Graph Understanding
                  and Question Answering},
  booktitle    = {NeurIPS},
  year         = {2024}
}

@article{DBLP:journals/ijcv/KrishnaZGJHKCKL17,
  author       = {Ranjay Krishna and
                  Yuke Zhu and
                  Oliver Groth and
                  Justin Johnson and
                  Kenji Hata and
                  Joshua Kravitz and
                  Stephanie Chen and
                  Yannis Kalantidis and
                  Li{-}Jia Li and
                  David A. Shamma and
                  Michael S. Bernstein and
                  Li Fei{-}Fei},
  title        = {Visual Genome: Connecting Language and Vision Using Crowdsourced Dense
                  Image Annotations},
  journal      = {IJCV},
  year         = {2017}
}

@inproceedings{DBLP:conf/emnlp/WangC0PWL00LLLZ22,
  author       = {Xiaozhi Wang and
                  Yulin Chen and
                  Ning Ding and
                  Hao Peng and
                  Zimu Wang and
                  Yankai Lin and
                  Xu Han and
                  Lei Hou and
                  Juanzi Li and
                  Zhiyuan Liu and
                  Peng Li and
                  Jie Zhou},
  title        = {{MAVEN-ERE:} {A} Unified Large-scale Dataset for Event Coreference,
                  Temporal, Causal, and Subevent Relation Extraction},
  booktitle    = {EMNLP},
  year         = {2022}
}

@misc{qwen3technicalreport,
      title={Qwen3 Technical Report}, 
      author={Qwen Team},
      year={2025}
}

@article{DBLP:journals/corr/abs-2309-12892,
  author       = {Zhilei Hu and
                  Zixuan Li and
                  Daozhu Xu and
                  Long Bai and
                  Cheng Jin and
                  Xiaolong Jin and
                  Jiafeng Guo and
                  Xueqi Cheng},
  title        = {ProtoEM: {A} Prototype-Enhanced Matching Framework for Event Relation
                  Extraction},
  journal      = {CoRR},
  year         = {2023}
}

@inproceedings{DBLP:conf/icml/StandleyZCGMS20,
  author       = {Trevor Standley and
                  Amir Zamir and
                  Dawn Chen and
                  Leonidas J. Guibas and
                  Jitendra Malik and
                  Silvio Savarese},
  title        = {Which Tasks Should Be Learned Together in Multi-task Learning?},
  booktitle    = {ICML},
  year         = {2020}
}

@article{DBLP:journals/corr/Ruder17a,
  author       = {Sebastian Ruder},
  title        = {An Overview of Multi-Task Learning in Deep Neural Networks},
  journal      = {CoRR},
  year         = {2017}
}

@inproceedings{DBLP:conf/icml/LaffertyMP01,
  author       = {John D. Lafferty and
                  Andrew McCallum and
                  Fernando C. N. Pereira},
  title        = {Conditional Random Fields: Probabilistic Models for Segmenting and
                  Labeling Sequence Data},
  booktitle    = {ICML},
  year         = {2001}
}

@inproceedings{DBLP:conf/cvpr/ZellersYTC18,
  author       = {Rowan Zellers and
                  Mark Yatskar and
                  Sam Thomson and
                  Yejin Choi},
  title        = {Neural Motifs: Scene Graph Parsing With Global Context},
  booktitle    = {CVPR},
  year         = {2018}
}

@inproceedings{DBLP:conf/nips/HamiltonYL17,
  author       = {William L. Hamilton and
                  Zhitao Ying and
                  Jure Leskovec},
  title        = {Inductive Representation Learning on Large Graphs},
  booktitle    = {NeurIPS},
  year         = {2017}
}

@article{DBLP:journals/corr/abs-1806-01261,
  author       = {Peter W. Battaglia and
                  Jessica B. Hamrick and
                  Victor Bapst and
                  Alvaro Sanchez{-}Gonzalez and
                  Vin{\'{\i}}cius Flores Zambaldi and
                  Mateusz Malinowski and
                  Andrea Tacchetti and
                  David Raposo and
                  Adam Santoro and
                  Ryan Faulkner and
                  {\c{C}}aglar G{\"{u}}l{\c{c}}ehre and
                  H. Francis Song and
                  Andrew J. Ballard and
                  Justin Gilmer and
                  George E. Dahl and
                  Ashish Vaswani and
                  Kelsey R. Allen and
                  Charles Nash and
                  Victoria Langston and
                  Chris Dyer and
                  Nicolas Heess and
                  Daan Wierstra and
                  Pushmeet Kohli and
                  Matthew M. Botvinick and
                  Oriol Vinyals and
                  Yujia Li and
                  Razvan Pascanu},
  title        = {Relational inductive biases, deep learning, and graph networks},
  journal      = {CoRR},
  year         = {2018}
}

@article{wang2025generative,
	title={Generative graph pattern machine},
	author={Wang, Zehong and Zhang, Zheyuan and Ma, Tianyi and Zhang, Chuxu and Ye, Yanfang},
	journal={NeurIPS},
	year={2026}
}

@article{sanh2021multitask,
  title={Multitask prompted training enables zero-shot task generalization},
  author={Sanh, Victor and Webson, Albert and Raffel, Colin and Bach, Stephen H and Sutawika, Lintang and Alyafeai, Zaid and Chaffin, Antoine and Stiegler, Arnaud and Scao, Teven Le and Raja, Arun and others},
  journal={CoRR},
  year={2021}
}

@article{zhang2026instruction,
  title={Instruction tuning for large language models: A survey},
  author={Zhang, Shengyu and Dong, Linfeng and Li, Xiaoya and Zhang, Sen and Sun, Xiaofei and Wang, Shuhe and Li, Jiwei and Hu, Runyi and Zhang, Tianwei and Wang, Guoyin and others},
  journal={ACM Comput. Surv.},
  year={2026}
}

@article{khashabi2020unifiedqa,
  title={Unifiedqa: Crossing format boundaries with a single qa system},
  author={Khashabi, Daniel and Min, Sewon and Khot, Tushar and Sabharwal, Ashish and Tafjord, Oyvind and Clark, Peter and Hajishirzi, Hannaneh},
  journal={CoRR},
  year={2020}
}

@inproceedings{mishra2022cross,
  title={Cross-task generalization via natural language crowdsourcing instructions},
  author={Mishra, Swaroop and Khashabi, Daniel and Baral, Chitta and Hajishirzi, Hannaneh},
  booktitle={ACL},
  year={2022}
}

@article{lu2019vilbert,
  title={Vilbert: Pretraining task-agnostic visiolinguistic representations for vision-and-language tasks},
  author={Lu, Jiasen and Batra, Dhruv and Parikh, Devi and Lee, Stefan},
  journal={NeurIPS},
  year={2019}
}

@article{tan2019lxmert,
  title={Lxmert: Learning cross-modality encoder representations from transformers},
  author={Tan, Hao and Bansal, Mohit},
  journal={CoRR},
  year={2019}
}

@inproceedings{chen2020uniter,
  title={Uniter: Universal image-text representation learning},
  author={Chen, Yen-Chun and Li, Linjie and Yu, Licheng and El Kholy, Ahmed and Ahmed, Faisal and Gan, Zhe and Cheng, Yu and Liu, Jingjing},
  booktitle={ECCV},
  year={2020}
}

@inproceedings{zhang2022look,
  title={Look twice as much as you say: Scene graph contrastive learning for self-supervised image caption generation},
  author={Zhang, Chunhui and Huang, Chao and Li, Youhuan and Zhang, Xiangliang and Ye, Yanfang and Zhang, Chuxu},
  booktitle={CIKM},
  year={2022}
}

@inproceedings{zhang2019heterogeneous,
  title={Heterogeneous graph neural network},
  author={Zhang, Chuxu and Song, Dongjin and Huang, Chao and Swami, Ananthram and Chawla, Nitesh V},
  booktitle={KDD},
  year={2019}
}

@inproceedings{guo2021few,
  title={Few-shot graph learning for molecular property prediction},
  author={Guo, Zhichun and Zhang, Chuxu and Yu, Wenhao and Herr, John and Wiest, Olaf and Jiang, Meng and Chawla, Nitesh V},
  booktitle={WWW},
  year={2021}
}

@inproceedings{ju2022grape,
  title={Grape: Knowledge Graph Enhanced Passage Reader for Open-domain Question Answering},
  author={Ju, Mingxuan and Yu, Wenhao and Zhao, Tong and Zhang, Chuxu and Ye, Yanfang},
  booktitle={EMNLP},
  year={2022}
}

@inproceedings{guo2020graseq,
  title={GraSeq: graph and sequence fusion learning for molecular property prediction},
  author={Guo, Zhichun and Yu, Wenhao and Zhang, Chuxu and Jiang, Meng and Chawla, Nitesh V},
  booktitle={CIKM},
  year={2020}
}

@article{liinstance,
  title={Instance-Aware Graph Prompt Learning},
  author={Li, Jiazheng and Li, Jundong and Zhang, Chuxu},
  journal={TMLR},
  year={2025}
}

@inproceedings{wang2025can,
  title={Can LLMs convert graphs to text-attributed graphs?},
  author={Wang, Zehong and Liu, Sidney and Zhang, Zheyuan and Ma, Tianyi and Zhang, Chuxu and Ye, Yanfang},
  booktitle={NAACL},
  year={2025}
}

@article{thapaliya2025semantic,
  title={Semantic Refinement with LLMs for Graph Representations},
  author={Thapaliya, Safal and Wang, Zehong and Li, Jiazheng and Li, Ziming and Ye, Yanfang and Zhang, Chuxu},
  journal={arXiv preprint arXiv:2512.21106},
  year={2025}
}

@inproceedings{jumulti,
  title={Multi-task Self-supervised Graph Neural Networks Enable Stronger Task Generalization},
  author={Ju, Mingxuan and Zhao, Tong and Wen, Qianlong and Yu, Wenhao and Shah, Neil and Ye, Yanfang and Zhang, Chuxu},
  booktitle={ICLR},
  year={2023}
}

@inproceedings{ma2025llm,
  title={Llm-empowered class imbalanced graph prompt learning for online drug trafficking detection},
  author={Ma, Tianyi and Qian, Yiyue and Wang, Zehong and Zhang, Zheyuan and Zhang, Chuxu and Ye, Yanfang},
  booktitle={ACL},
  year={2025}
}

@inproceedings{chen2018gradnorm,
	author       = {Zhao Chen and
	Vijay Badrinarayanan and
	Chen{-}Yu Lee and
	Andrew Rabinovich},
	title        = {GradNorm: Gradient Normalization for Adaptive Loss Balancing in Deep
	Multitask Networks},
	booktitle    = {ICML},
	year         = {2018}
}

@inproceedings{DBLP:conf/icml/00120C000Z25,
  author       = {Jiazheng Li and
                  Lu Yu and
                  Qing Cui and
                  Zhiqiang Zhang and
                  Jun Zhou and
                  Yanfang Ye and
                  Chuxu Zhang},
  title        = {{MASS:} Mathematical Data Selection via Skill Graphs for Pretraining
                  Large Language Models},
  booktitle    = {ICML},
  year         = {2025}
}

@inproceedings{DBLP:conf/icml/WangZMCZ025,
  author       = {Zehong Wang and
                  Zheyuan Zhang and
                  Tianyi Ma and
                  Nitesh V. Chawla and
                  Chuxu Zhang and
                  Yanfang Ye},
  title        = {Towards Graph Foundation Models: Learning Generalities Across Graphs
                  via Task-Trees},
  booktitle    = {ICML},
  year         = {2025}
}

@inproceedings{DBLP:conf/kdd/ZhangSHSC19,
  author       = {Chuxu Zhang and
                  Dongjin Song and
                  Chao Huang and
                  Ananthram Swami and
                  Nitesh V. Chawla},
  title        = {Heterogeneous Graph Neural Network},
  booktitle    = {KDD},
  year         = {2019}
}

@inproceedings{DBLP:conf/ijcai/LiLLLZ25,
  author       = {Ziming Li and
                  Youhuan Li and
                  Yuyu Luo and
                  Guoliang Li and
                  Chuxu Zhang},
  title        = {Graph Neural Networks for Databases: {A} Survey},
  booktitle    = {IJCAI},
  year         = {2025}
}

@article{ma2025sarvlm,
  title={SARVLM: A vision language foundation model for semantic understanding and target recognition in SAR imagery},
  author={Ma, Qiwei and Wang, Zhiyu and Liu, Wang and Lu, Xukun and Deng, Bin and Duan, Puhong and Kang, Xudong and Li, Shutao},
  journal={CoRR},
  year={2025}
}

@inproceedings{DBLP:conf/iclr/VelickovicCCRLB18,
  author       = {Petar Velickovic and
                  Guillem Cucurull and
                  Arantxa Casanova and
                  Adriana Romero and
                  Pietro Li{\`{o}} and
                  Yoshua Bengio},
  title        = {Graph Attention Networks},
  booktitle    = {ICLR},
  year         = {2018}
}

@inproceedings{DBLP:conf/iclr/KipfW17,
  author       = {Thomas N. Kipf and
                  Max Welling},
  title        = {Semi-Supervised Classification with Graph Convolutional Networks},
  booktitle    = {ICLR},
  year         = {2017}
}

@inproceedings{DBLP:conf/esws/SchlichtkrullKB18,
  author       = {Michael Sejr Schlichtkrull and
                  Thomas N. Kipf and
                  Peter Bloem and
                  Rianne van den Berg and
                  Ivan Titov and
                  Max Welling},
  title        = {Modeling Relational Data with Graph Convolutional Networks},
  booktitle    = {ESWC},
  year         = {2018}
}
